\pdfoutput=1

\documentclass[11pt]{article}

\usepackage[]{acl}

\usepackage{times}
\usepackage{latexsym}

\usepackage[T1]{fontenc}

\usepackage[utf8]{inputenc}
\usepackage{amssymb}
\usepackage{microtype}

\usepackage{stmaryrd}
\usepackage{inconsolata}
\usepackage{microtype}
\usepackage{enumitem}
\usepackage{inconsolata}
\usepackage{todonotes}
\usepackage[normalem]{ulem}
\usepackage{booktabs}
\usepackage{subcaption}
\usepackage{hyperref}
\usepackage{url}
\usepackage{graphicx}
\usepackage{multirow}
\usepackage{arabtex}
\usepackage{utf8}
\usepackage{comment}

\usepackage[utf8]{inputenc}
\setcode{utf8}
\usepackage{listings}
\usepackage{color}

\usepackage{pifont}
\newcommand{\cmark}{\ding{51}}%
\newcommand{\xmark}{\ding{55}}%

\definecolor{blue}{rgb}{0,0, 0.6}
\definecolor{dkgreen}{rgb}{0,0.6,0}
\definecolor{gray}{rgb}{0.5,0.5,0.5}
\definecolor{mauve}{rgb}{0.58,0,0.82}
\definecolor{mauve}{rgb}{0,0,0}
\definecolor{black}{rgb}{0,0,0}

\usepackage[normalem]{ulem}

\usepackage{soul}
\definecolor{tri}{rgb}{.25,.88,.82}
\definecolor{lilac}{rgb}{0.85,0.64,0.85}
\definecolor{atomictangerine}{rgb}{1.0, 0.6, 0.4}
\definecolor{lightblue}{rgb}{0.53, 0.81, 0.98} 

\newcommand{\arabench}{LAraBench}

\title{\arabench: Benchmarking Arabic AI with Large Language Models}

\author{
    Ahmed Abdelali,\textsuperscript{$\dagger$}\thanks{~~The contribution was made while the author was at the Qatar Computing Research Institute.}\textsuperscript{~~$^1$}
    Hamdy Mubarak\thanks{~~Equal contribution.},\textsuperscript{$^1$}
    Shammur Absar Chowdhury,\textsuperscript{$^1$}
    Maram Hasanain,\textsuperscript{$^1$} \\
    {\bf
        Basel Mousi,\textsuperscript{$^1$}
        Sabri Boughorbel,\textsuperscript{$^1$}
        Samir Abdaljalil,\textsuperscript{$^1$}
        Yassine El Kheir,\textsuperscript{$^1$}
        Daniel Izham,\textsuperscript{$^2$} 
    }\\
    {\bf
        Fahim Dalvi,\textsuperscript{$^1$}
        Majd Hawasly,\textsuperscript{$^1$}
        Nizi Nazar,\textsuperscript{$^1$}
        Yousseif Elshahawy,\textsuperscript{$^2$} 
    }\\
    {\bf
        Ahmed Ali,\textsuperscript{$^1$}
        Nadir Durrani,\textsuperscript{$^1$}
        Natasa Milic-Frayling,\textsuperscript{$^1$}
        Firoj Alam\textsuperscript{$^1$}
    }\\
    \textsuperscript{$^1$}Qatar Computing Research Institute, HBKU, Qatar, 
    \textsuperscript{$^2$}Kanari AI, Doha, Qatar \\
    fialam@hbku.edu.qa \\
}

\begin{document}
\maketitle
\begin{abstract}

Recent advancements in Large Language Models (LLMs) have significantly influenced the landscape of language and speech research. Despite this progress, these models lack specific benchmarking against state-of-the-art (SOTA) models tailored to particular languages and tasks. \textit{\arabench{}} addresses this gap for Arabic Natural Language Processing (NLP) and Speech Processing tasks, including sequence tagging and content classification across different domains. We utilized models such as GPT-3.5-turbo, GPT-4, BLOOMZ, Jais-13b-chat, Whisper, and USM, employing zero and few-shot learning techniques to tackle 33 distinct tasks across 61 publicly available datasets. This involved 98 experimental setups, encompassing $\sim$296K data points, $\sim$46 hours of speech, and 30 sentences for Text-to-Speech (TTS). This effort resulted in 330+ sets of experiments. Our analysis focused on measuring the performance gap between SOTA models and LLMs. The overarching trend observed was that SOTA models generally outperformed LLMs in zero-shot learning, with a few exceptions. Notably, larger computational models with few-shot learning techniques managed to reduce these performance gaps. Our findings provide valuable insights into the applicability of LLMs for Arabic NLP and speech processing tasks. 

\end{list} 
\end{abstract}

\section{Introduction}
\label{sec:introduction}


Generative Pre-trained Transformer (GPT) models are examples of large language models (LLMs)\footnote{We are referring to models with billions of parameters as LLMs.} trained on massive datasets and using hundreds of millions of parameters. Several LLMs have been recently released for use through APIs or pre-trained models 
and have demonstrated a high level of coherence in generating content in response to specific user tasks. However, quality assessments of released LLMs generally lack references to previous research and comparison with state-of-the-art (SOTA) methods that the research community has used for systematic evaluation and monitoring of scientific progress for various languages and tasks. 

\begin{figure}[]
    \centering
\includegraphics[width=\columnwidth]{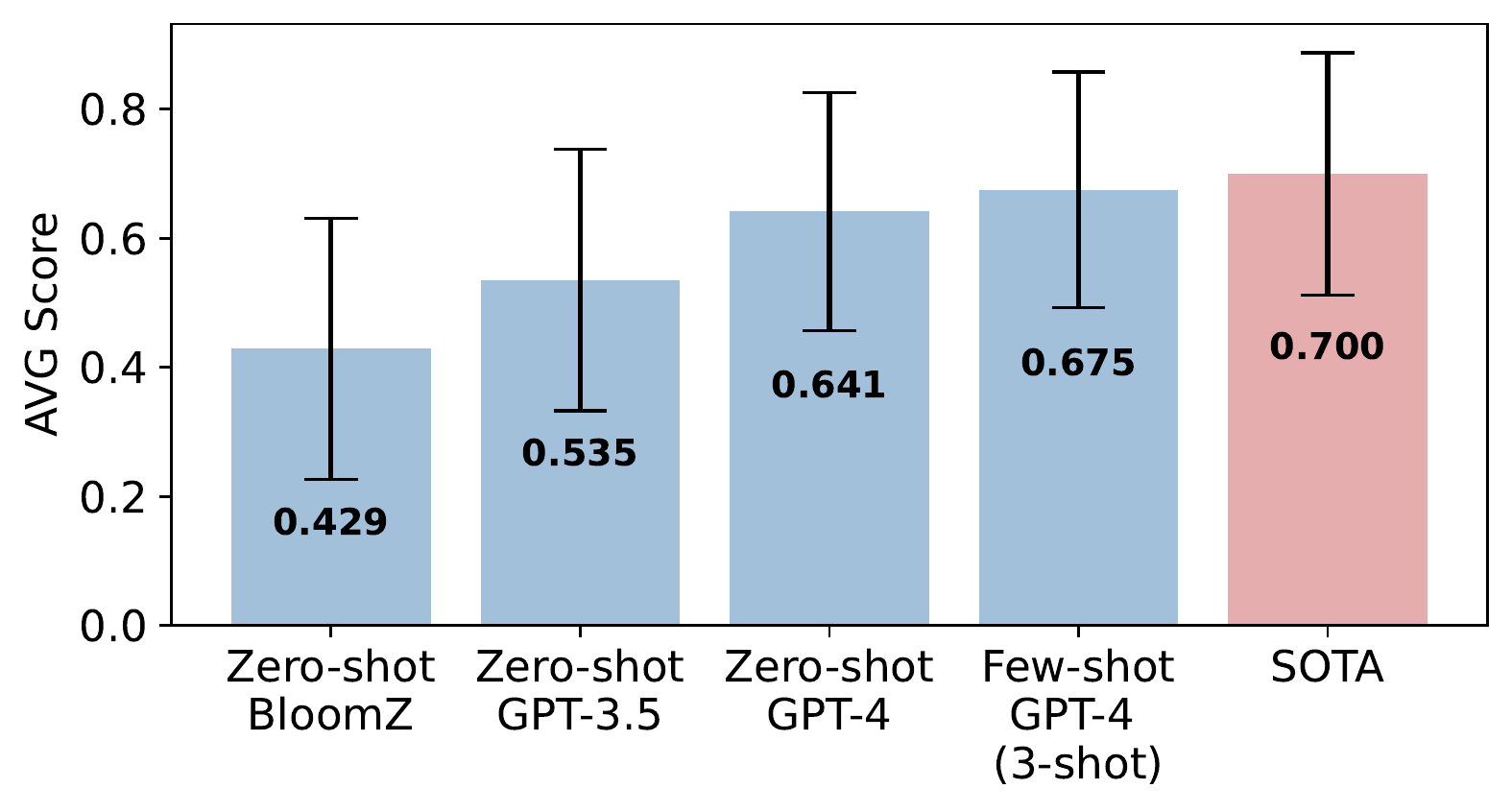}
    \caption{Average performance of the models as compared to SOTA across 21 unique NLP tasks and 31 testing setups.}.
    \vspace{-0.5cm}
    \label{fig:model_comparison}
    \vspace{-0.4cm}
\end{figure}


Several research initiatives have evaluated these large models' performance on standard NLP and speech processing tasks. The HELM project \citep{liang2022holistic} assessed English LLMs across various metrics and scenarios. BIG-Bench \citep{srivastava2022beyond} introduced a large-scale evaluation with 214 tasks, including low-resource languages. GPT2.5~\cite{radford2019language}, ChatGPT~\cite{openai2023gpt4}, and BLOOM~\cite{scao2022bloom}, 
were recently evaluated by \citet{bang2023multitask,ahuja2023mega,hendy2023good,tawkat2023gptaraeval}. Large speech models such as Whisper \cite{radford2022robust} and Universal Speech Model (USM) \cite{zhang2023google} were also explored for speech recognition and translation tasks. Initiatives such as SUPERB \cite{yang2021superb} were introduced to support benchmarking tools and leaderboards for several speech-related tasks. \citet{bubeck2023sparks} explored GPT-4's capabilities to determine if it surpasses mere memorization, possessing a profound and adaptable comprehension of concepts, skills, and domains. Their results indicate that GPT-4 demonstrates a higher level of general intelligence compared to its predecessors. 

\textit{\arabench{}} study fulfills an important objective of assessing the LLMs capabilities for supporting Arabic language processing tasks, for Modern Standard Arabic (MSA) and dialectal Arabic (DA), at the same level of depth and breadth as for English tasks. 
Our evaluation involves 61 publicly available datasets and 98 test setups used to perform and evaluate language processing tasks in both MSA and dialectal Arabic across various genres (e.g., news articles, tweets, meetings, telephony, and broadcast content). Our evaluation focuses on assessing the capabilities of GPT-3.5-turbo, GPT-4, Jais-13b-chat~\cite{sengupta2023jais}\footnote{We benchmarked Jais model on seven datasets using a zero-shot setting.} and BLOOMZ (176B) for NLP tasks, and of Whisper (Large, 1.55B) and USM (2B) for Speech processing, in both zero and few-shot settings. We investigate: 
\textit{\textbf{(i)} can LLMs effectively perform Arabic NLP and Speech processing tasks without prior task-specific knowledge (zero-shot)?} \textit{\textbf{(ii)} how does performance vary across tasks with different complexities in zero- and few-shot settings?}
\textit{\textbf{(iii)} how do LLMs compare to current SOTA models, and are open LLMs as effective as the commercially available (closed) models?}
\noindent Our investigation reveals unique insights about LLMs' performance on Arabic NLP and Speech tasks:
\begin{description}[leftmargin=*]
    \item [A. Zero-shot Multi-task Performer.] GPT-4 outperforms other models 
    in majority of the NLP tasks (see Figure \ref{fig:model_comparison}). However, a large performance gap between GPT-4 and SOTA models  remains due to the higher quality SOTA models. For speech tasks, USM outperforms all the Whisper variants and performs comparably with SOTA.
    \item [B. Few-shot and SOTA.] GPT-4 reduces the performance gap with SOTA in the few-shot (only 3-shots) setting (see Figure~\ref{fig:model_comparison}). This significant performance gain is noticed for almost all tasks, particularly for more complex semantic and question-answering tasks compared to syntactic and segmentation tasks. 
    Similarly, Whisper models exhibit promising results in speech recognition with just 2 hours of speech examples in few-shot finetuning. Open models (BLOOMZ and Whisper) performed poorly w.r.t. to their commercially available counterparts. However, fine-tuning with more instructions may help these open models to achieve closer performance to SOTA and other closed LLMs. 
    
    \item [C. MSA {\em vs }Dialect.] The gaps in LLMs' performance between MSA and dialectal datasets (e.g., for machine translation (MT) and speech recognition task) are more pronounced, indicating ineffectiveness of LLMs for under-represented  dialects. 
    \item [D. Hallucination and Data Contamination.] GPT models, specially GPT-3.5, suffer from the hallucination problem. We noticed the model insert extra information (e.g., for MT task with Bible dataset) from its parametric memory. 
Benchmarking LLMs raises concerns about their exposure to existing datasets. In our study, we utilized datasets that were released after the cut-off date of GPT's training (September 2021). Moreover, we applied a prompt-based method with guided instructions \cite{golchin2023time} on nine datasets using GPT-4, to determine if datasets are contaminated. Our experiments revealed that GPT-4 could not produce any examples from these datasets. 

\end{description}



To the best of our knowledge, \textit{LAraBench} is the \textit{first} comprehensive effort that includes commercial (close) and open source LLMs and evaluates zero- and few-shot settings for a wide range of Arabic NLP and Speech tasks. It is the \textit{first} to include the evaluation of Whisper and USM models for Arabic ASR and the \textit{first} to report benchmarks for a standard Arabic TTS generative model. All resources and findings of the \textit{LAraBench} study made publicly available 
to scale up the effort through our LLMeBench framework \cite{dalvi-etal-2024-llmebench}.\footnote{\url{https://github.com/qcri/LLMeBench}}

\section{Tasks and Datasets}
\label{sec:tasks_and_dataset}

The \arabench{} study was designed with an ambitious goal of empowering the research community and practitioners with the most comprehensive evaluation of LLMs for Arabic NLP and Speech tasks to date.
It includes 61 publicly available datasets to support 9 task groups\footnote{
Our task categorization is derived from the taxonomy outlined in the list of tracks established by ACL 2022.} discussed below.
We briefly describe each task and refer to Appendix~\ref{sec:appx-tasks-datasets} for a comprehensive description of tasks and datasets.



\paragraph{Word Segmentation, Syntax and Information Extraction.}
We explore six sequence tagging tasks:  i) word segmentation, ii) POS-tagging, iii) lemmatization, iv) diacritization, v) parsing, and vi) named-entity recognition (NER), using publicly available datasets. 
We also include a dialect identification task (e.g., Egyptian dialect) since vocabulary, pronunciation, and idiomatic expressions vary across dialects. For our benchmarking we used QADI~\cite{abdelali-etal-2021-qadi} and ADI (in-house) datasets.


\paragraph{Machine Translation (MT).}
Machine translation of Arabic is challenging due to morphological complexity 
and dialectal variations \cite{DurraniAI14,sajjad-etal-2017-challenging}. 
We experiment with AraBench \cite{sajjad2020arabench}, an extensive suite of data offering 4 coarse, 15 fine-grained and 25 city-level dialect categories. The dataset covers diverse genres such as media, chat, religion, and travel.




\paragraph{Sentiment, Stylistic and Emotion Analysis.}

These tasks involve understanding and analyzing aspects of human expression and communication. 
We benchmark sentiment analysis, emotion recognition, stance detection, and sarcasm detection with datasets from \citet{elmadany2018arsas}, \citet{mohammad2018semeval}, \citet{chouigui2017ant}, and \citet{abufarha-etal-2021-arsarcasm-v2}, respectively.

\paragraph{News Categorization.}
This task involves classification of news articles into pre-defined categories or topics \cite{sebastiani2002machine}. 
We benchmark news categorization task using SANAD news article corpus~\cite{einea2019sanad} and ASND social media dataset~\cite{chowdhury2020improving}.

\paragraph{Demographic Attributes.}
Demographic information, including gender, age, and country of origin, hold significant value across various applications such as population analysis. 
 We include datasets that enable experimentation with tasks of identifying country, gender~\cite{mubarak2022arabgend} and location \cite{mubarak2021ul2c}.


\paragraph{Ethics and NLP: Factuality, Disinformation and Harmful Content Detection.}
These tasks have emerged as critical areas within the field of NLP. We benchmark several \emph{detection} tasks, such as: 
i) offensive language~\cite{zampieri2020semeval},  
ii) hate speech~\cite{mubarak2021adult},  
iii) adult content~\cite{mubarak2021adult},  
iv) spam~\cite{mubarak2020spam}, 
v) subjectivity \cite{clef-checkthat:2023:task2}, 
vi) propaganda~\cite{alam2022overview}, 
vii) check-worthiness~\cite{nakov2022overview}, 
viii) factuality using the datasets~\citet{baly-etal-2018-predicting,alam-etal-2021-fighting-covid,khouja-2020-stance}, 
ix) claim~\cite{alam2020call2arms}, 
x) harmful content~\cite{nakov2022overview}, and
xi) attention-worthiness \citep{nakov2022clef}.


\paragraph{Semantics.}
This task group includes Semantic Textual Similarity (STS) and Natural Language Inference (NLI). 
We benchmark STS using two datasets: 
SemEval-2017 STS task~\cite{cer-etal-2017-semeval} and similarity in Arabic question pairs, as explored by \citet{seelawi2019nsurl}.
For the XNLI task, we used the translated version of Arabic from XNLI corpus~\cite{conneau2018xnli}. 

\paragraph{Question Answering (QA).}
For the QA task, we employed ARCD~\cite{mozannar2019neural}, MLQA~\cite{lewis2019mlqa}, TyDiQA~\cite{tydiqa}, and XQuAD~\cite{artetxe2019cross} datasets.

\paragraph{Speech Processing.} 
We evaluate the large speech models on two tasks: speech recognition (ASR) and text-to-speech (TTS) synthesis. For ASR, we include datasets from varying domains and dialects, e.g. MGB2~\cite{ali2016mgb}, QASR.CS~\cite{mubarak2021qasr} and ESCWA.CS~\cite{ali2021arabic}. For TTS, we evaluated with in-house 30 test sentences \cite{abdelali-2022-natiq}, covering diverse topics (e.g., education, health).




\section{Methodology}
\label{sec:benchmarking_methods}
For benchmarking of Arabic NLP and Speech processing tasks, we use zero- and few-shot learning involving GPT-3.5-Turbo, GPT-4, BLOOMZ and Jais-13b-chat for NLP, and Whisper (small, medium, and large), USM and Amazon Polly for Speech. We also compared LLM's performance with the respective SOTA models. 

The use and evaluation of LLMs involve prompting and post-processing of output to extract the expected content. Therefore, for each task, we explored a number of prompts, guided by the same instruction and format as recommended in the Azure OpenAI Studio Chat playground, and PromptSource~\cite{bach2022promptsource}. 
After obtaining a reasonable prompt,\footnote{Note that our objective was not to identify the optimal prompt but rather to find a prompt that would yield reasonable performance without incurring excessive costs.} we used it to complete the evaluation of the task using modality-specific API services, e.g., OpenAI API from Azure for NLP tasks and Google's USM API for Speech tasks. 
As for BLOOMZ and Jais-13b-chat, we use an on-premises hosted version.

We based our model selection on factors like performance, language support, and accessibility. For NLP tasks, we chose OpenAI models because they consistently outperformed others for English tasks. Initially, we used GPT-3.5 and later transitioned to GPT-4 when it became available. Limited budget and lack of Arabic language support led us to avoid other closed models. Among open models, we selected BLOOMZ because it's a large multilingual model, including 4\% Arabic content. 
Recently released Arabic-focused models, such as Jais~\cite{sengupta2023jais} and AceGPT~\cite{huang2023acegpt}, have become available as we carry out this study. In our experiments, we benchmarked the Jais-13b-chat model across datasets related to seven tasks. A comprehensive exploration of these newly released Arabic models will be incorporated into our future studies.
For ASR, we chose Whisper and USM due to their excellent performance in recent studies. 


\subsection{Models and Prompts for NLP Tasks}
\label{ssec:zero_few_shot}

\paragraph{Zero-shot Setup.}
For tasks with GPT-3.5-Turbo, GPT-4, BLOOMZ and Jais-13b-chat, we use zero-shot prompting giving natural language instructions describing the task and specify the expected output. 
Prompts allows LLMs to learn context and narrows the inference space to produces accurate output.




\paragraph{Few-shot Setup.}
In order to explore the maximum potential of specific LLMs, e.g., GPT-4 model, we used available training data to select few-shot examples and provide context for the task. For a few tasks and datasets (e.g., location, name to country), training sets are either private or not available and therefore they could not be included in our few-shot experiments. 
We used maximal marginal relevance-based (MMR) selection to construct example sets that are deemed relevant and diverse \cite{carbonell1998use}, following the proven method by~\citet{ye2022complementary}. The MMR method computes the similarity between a test example and the example pool (e.g., training dataset) and  selects $m$ examples (shots). We apply MMR on top of embeddings of multilingual sentence-transformers \cite{reimers2019sentence}. 
In our few-shot investigation, we performed experiments on all tasks and datasets using only 3-shots to primarily reduce computational and API expenses. Additionally, we expanded our analysis to include 3, 5, and 10 instances across seven distinct datasets drawn from various task categories.
More details are provided in Section \ref{ssec:appx_extended_few_shot_results} of the Appendix.



\paragraph{Prompts Design.}
Prompt design is a complex and iterative process that presents challenges due to the unknown representation of information within LLMs and a need for different types of outputs across tasks, e.g., token classification vs. sentence classification. The instructions expressed in our prompts were in English, including the content examples in Arabic. In Appendix \ref{sec:appx-prompts}, we provide examples of prompts for different tasks, which are also released with the LLMeBench framework.  
We also examined Arabic instructions in our study, to understand the effect of native language prompts. For this set of experiments we selected seven datasets from seven different task groups. More details can be found in Section \ref{ssec:appx_native_language_prompts} (Appendix).

\paragraph{Post Processing.} Outputs of LLMs are post-processed to enable automatic comparison with gold standard labels. Depending on the task, this may include mapping prefixes, or filtering tokens. For example, for POS tagging, the tags \textit{`preposition'}, \textit{`P'}, \textit{`PRP'}, \textit{`\<حرف جر>'}, are mapped onto  \textit{PREP}. For NER, the model switches the tag of the prediction i.e., B-PER predicts as PER-B, and therefore requires remapping of the NER tags.

\subsection{Models and Prompts for Speech Tasks}
\label{ssec:zero_few_shot_speech}
We use zero- and few-shot settings to benchmark large speech models. For ASR, we use three Whisper models (OpenAI) -- small, medium, and large, and the USM model (Google). For the details of the models, see Table~\ref{sec:appx-speech-model-parameters} in Appendix. We compare these large models to SOTA: supervised QCRI-KANARI\footnote{https://fenek.ai/} conformer-based \cite{chowdhury2021towards} offline and RNN-T based streaming ASR.\footnote{https://arabicasr.kanari.ai/} For the TTS task, we compare two public systems: Amazon Polly TTS engine\footnote{https://aws.amazon.com/polly/} and QCRI-KANARI (Q-K) TTS \cite{abdelali-2022-natiq} system.\footnote{https://arabictts.kanari.ai/} 


\paragraph{Zero-shot Setup.} For zero-shot setup, we use the initial (or pre-trained) weights of Whisper and API of USM models with a goal to benchmark the performances of these LLMs in different domains, for different Arabic dialects, and for code-switching with no domain knowledge. As a prompt to the model, we passed only a language flag.

\paragraph{Few-shot Setup.} 
Under this setup, we fine-tune Whisper (small and large) with 2 hours of domain-specific speech data and compare it with the SOTA models trained from scratch with 3K hours of speech.

\paragraph{ASR Post Processing.} 
ASR is evaluated based on word error rate (WER) that aligns the model's output with reference transcription and penalizes the output based on insertion, deletion, and substitution errors. The measure is unable to disambiguate code-switching and minor formatting differences introduced by multilingual scripts or non-standardized orthography. Hence, post-processing is a crucial component.
We normalized `alif', `ya' and ta-marbuta', and adapted a minimalist Global Mapping File (GLM) \cite{chowdhury2021towards} to transliterate common words and handle rendering mismatch. Thus keeping room for further improvement with more enhanced post-processing.

\subsection{Random Baseline}
\label{sec:random_baseline}
We also calculated a random baseline for the 
NLP tasks (further details in Appendix, Section \ref{ssec:appx_random_baseline}). The aim is to determine if the LLMs predictions are not merely the result of chance. It also serves as a lower limit to be expected for each task. 

\subsection{SOTA Models}
Our study benchmarks large language models (LLMs) drawing comparisons against a wide variety of methods employed in various studies. These encompass state-of-the-art results utilizing diverse architectures, including LSTM, CRF, GRU, SVM, and various Arabic and multilingual transformer models such as AraBERT, XLM-r, and mT5 etc.


\subsection{Evaluation Metrics}
\label{sec:evaluation_metric}
To measure the performance of each task, we followed current state-of-art references and used the metric reported in the respective work. This includes: Accuracy (Acc), F1 (macro, micro, and weighted), word error rate (WER), Jaccard Similarity (JS), Pearson Correlation (PC), 
and mean opinion score (MOS) for naturalness, intelligibility and diacritization. We report average MOS (10-point Likert scale) from 3 native-annotators.
 



\section{Results and Discussion}
\label{sec:results_and_discussion}

In Tables \ref{tab:exp_results_nlp_tasks}, \ref{tab:exp_results_mt_only}, \ref{tab:exp_results_asr_only} and \ref{tab:exp_results_tts_only}, we report the results of different NLP and Speech related tasks. In the below sections, we summarize the results and challenges specific to the task groups.

\subsection{NLP Tasks}
In Table \ref{tab:exp_results_nlp_tasks}, we report the random baseline, GPT-3.5, GPT-4 (zero-shot and few-shot), and BLOOMZ and compare them to SOTA.\footnote{Note that some results are missing either due to the unavailability of training data required for few-shot experiments (marked with NA) or the incapability of the BLOOMZ model (marked with $\ddag$).} 
In almost all tasks, models outperform random baseline, indicating that the predictions of the models are not by chance. 
In the case of syntactic tasks such as segmentation, lemmatization, diacritization, POS, and NER, \emph{BLOOMZ} consistently failed to generate the desired output. This suggests a potential lack of understanding of the tasks at hand. Notably, in the diacritization task, the model failed to produce any diacritized content as instructed, instead returning a portion of the input. While the issue may be specific to the Arabic language, it merits investigation to determine if similar challenges exist in languages employing accented letters.


\begin{table*}[h!] 
\centering
\setlength{\tabcolsep}{4pt}
\resizebox{0.95\textwidth}{!}{
\begin{tabular}{lllllllll}
\toprule
\textbf{Task Name} & \textbf{Dataset}  & \textbf{Metric} & \textbf{\begin{tabular}[c]{@{}l@{}}Random \\ Baseline\end{tabular}}  & \textbf{BLOOMZ} & \textbf{\begin{tabular}[c]{@{}l@{}}Zero-shot \\ GPT-3.5\end{tabular}} & \textbf{\begin{tabular}[c]{@{}l@{}}Zero-shot \\ GPT-4\end{tabular}} & \textbf{\begin{tabular}[c]{@{}l@{}}Few-Shot\\ GPT-4\\ (3-shot)\end{tabular}} & \textbf{SOTA} \\ \midrule
\multicolumn{9}{c}{\textbf{Word Segmentation, Syntax and Information Extraction}} \\ \midrule
Segmentation & WikiNews & Acc & 0.272 & $\ddag$ & 0.195 & 0.252  & 0.927 & \textbf{0.990} \cite{abdelali2016farasa} \\
Segmentation & \citet{samih2017learning} & Acc$_{AVG}$  & 0.309 & $\ddag$ & 0.283 & 0.372  & 0.850 & \textbf{0.931} \citet{samih2017learning} \\
Lemmatization  & WikiNews & Acc & 0.348 & $\ddag$ & 0.471 & 0.397  & NA & \textbf{0.973} \cite{mubarak-2018-build}  \\
Diacritization & WikiNews & WER & 0.963 & $\ddag$ & 0.308 & 0.420  & 0.237 & \textbf{0.045} \cite{mubarak2019highly} \\
Diacritization & \citet{darwish2018diacritization} & WER & 0.999 & $\ddag$ & 0.928 & 0.899  & 0.994 & \textbf{0.031} \cite{darwish2018diacritization} \\
POS & WikiNews & Acc & 0.030 & $\ddag$ & 0.231 & 0.479  & 0.367 & \textbf{0.953} \cite{darwish2017arabic} \\
POS & \citet{samih2017learning} & Acc & 0.036 & $\ddag$  & 0.073 & 0.511  & 0.323 & \textbf{0.892} \citet{samih2017learning} \\
POS & GLUE (Arabic)   & Acc & 0.032 & $\ddag$  & 0.159 & 0.402  & 0.524 & \textbf{0.686} \cite{liang2020xglue} \\
Parsing & Conll2006 & UAS & 0.001 & $\ddag$ & 0.239 & 0.504  & 0.551 & \textbf{0.796} \cite{lei2014low} \\
NER & ANERcorp & F1$_{Macro}$   & 0.008 & $\ddag$ & 0.210 & 0.355  & 0.420 & \textbf{0.886} \cite{gridach2018deep}   \\
NER & Aqmar & F1$_{Macro}$   & 0.007 & $\ddag$ & 0.230 & 0.365  & 0.390 & \textbf{0.690} \cite{schneider2012coarse}  \\
NER & QASR & F1$_{Macro}$   & 0.009 & $\ddag$  & 0.208 & 0.504  & NA & \textbf{0.698} \cite{mubarak2021qasr}   \\
Dialect & QADI & F1$_{Macro}$   & 0.052 & 0.067 & 0.149 & 0.243  & NA & \textbf{0.600} \cite{abdelali-etal-2021-qadi}   \\
Dialect & ADI  & F1$_{Macro}$   & 0.092 & 0.098 & 0.169 & 0.229  & 0.260  & \textbf{0.26/0.57} (lexical/acoustic) (In-house) \\
\midrule
\multicolumn{9}{c}{\textbf{Sentiment, Stylistic and Emotion Analysis}}\\ \midrule
Sentiment  & ArSAS & F1$_{Macro}$   & 0.222 & 0.251   & 0.550 & 0.569  & 0.598 & \textbf{0.758} \cite{hassan2021asad}    \\
Emotion & SemEval18-Task1  & JS & 0.167 & 0.142   & 0.395 & 0.373  & 0.489 & \textbf{0.541} \cite{hassan2022cross}   \\
Stance & Unified-FC & F1$_{Macro}$   & 0.193 & 0.235   & 0.232 & 0.495  & 0.358 & \textbf{0.558} \cite{baly2018integrating} \\
Stance & ANS & F1$_{Macro}$   & 0.281 & 0.223   & 0.620 & 0.762  & 0.721 & \textbf{0.767} \cite{khouja-2020-stance} \\
Sarcasm & ArSarcasm & F1$_{(POS)}$   & 0.240 & 0.286 & 0.465 & 0.400  & \textbf{0.504}  & 0.460 \cite{farha2020arabic} \\
Sarcasm & ArSarcasm-2  & F1$_{(POS)}$   & 0.333 & 0.436 & 0.537 & 0.573  & 0.537 & \textbf{0.623} \cite{alharbi2021multi}  \\
\midrule
\multicolumn{9}{c}{\textbf{News Categorization}} \\ \midrule
News Cat.  & ASND & F1$_{Macro}$   & 0.048 & 0.371   & 0.512 & 0.667  & 0.594 & \textbf{0.770} \cite{chowdhury2020improving}    \\
News Cat.  & SANAD/Akhbarona  & Acc & 0.142 & 0.582   & 0.730 & 0.877  & 0.892 & \textbf{0.940} \cite{elnagar2020arabic} \\
News Cat.  & SANAD/AlArabiya  & Acc & 0.144 & 0.716   & 0.922 & 0.921  & 0.925 & \textbf{0.974} \cite{elnagar2020arabic}  \\
News Cat.  & SANAD/AlKhaleej  & Acc & 0.142 & 0.738   & 0.864 & 0.911  & 0.899 & \textbf{0.969} \cite{elnagar2020arabic}  \\
\midrule
\multicolumn{9}{c}{\textbf{Demographic Attributes}}\\ \midrule
Name Info  & ASAD & F1$_{Weighted}$ & 0.014 & $\ddag$ & 0.570 & \textbf{0.629} & NA & 0.530 (Under review)   \\
Location   & UL2C & F1$_{Macro}$   & 0.027 & 0.118 & 0.339 & 0.735  & NA & \textbf{0.881} \cite{mubarak2021ul2c}   \\
Gender & Arap-Tweet & F1$_{Macro}$   & 0.521 &  0.532   & 0.883 & 0.868  & \textbf{0.980}  & 0.821 \cite{zaghouani-charfi-2018-arap} \\ 
\midrule
\multicolumn{9}{c}{\textbf{Ethics and NLP: Factuality, Disinformation and Harmful Content Detection}}\\ \midrule
Offensive lang. & OffensEval2020  & F1$_{Macro}$   & 0.454 &  0.533   & 0.460 & 0.623  & 0.874 & \textbf{0.905} \cite{mubarak-etal-2020-overview} \\
Hate Speech & OSACT2020 & F1$_{Macro}$   & 0.376 & 0.503   & 0.430 & 0.669  & 0.644 & \textbf{0.823} \cite{mubarak-etal-2020-overview} \\
Adult Content  & ASAD & F1$_{Macro}$   & 0.421 &  0.513 & 0.460 & 0.727  & 0.832 & \textbf{0.889} \cite{mubarak2021adult}  \\
Spam & ASAD & F1$_{Macro}$   & 0.405 & 0.152 & 0.440 & 0.745  & NA & \textbf{0.989} \cite{hassan2021asad}    \\
Subjectivity & In-house & F1$_{Macro}$   & 0.496 & 0.428   & 0.670 & 0.677  & \textbf{0.745}  & 0.730 (In-house)\\
Propaganda & WANLP22  & F1$_{Micro}$   & 0.139 & 0.108   & 0.353 & 0.472  & 0.537 & \textbf{0.649} \cite{samir2022ngu_cnlp} \\
Check-worthy & CT–CWT–22 & F1$_{(POS)}$   & 0.398 & 0.431   & 0.526 & 0.560  & 0.554 & \textbf{0.628} \cite{du2022nus}  \\
Factuality & COVID-19 Disinfo. & F1$_{Weighted}$ & 0.582 & \textbf{0.749} & 0.393 & 0.485  & 0.491 & \textbf{0.831} \cite{alam-etal-2021-fighting-covid}    \\
Factuality & Unified-FC & F1$_{Macro}$   & 0.464 & 0.460   & 0.306 & 0.581  & \textbf{0.621}  & \(\varnothing\)   \\
Factuality & ANS   & F1$_{Macro}$   & 0.505  & 0.550   & 0.252 & 0.539  & \textbf{0.704}  & 0.643 \cite{khouja-2020-stance} \\
Claim  & CT–CWT–22 & Acc & 0.498  & 0.532   & \textbf{0.703} & 0.587  & 0.686 & 0.570 \cite{clef-checkthat:2022:task1:Kutlu}    \\
Harmful content & CT–CWT–22 & F1$_{(POS)}$   & 0.269 & 0.144   & 0.471 & 0.533  & 0.494 & \textbf{0.557} \cite{clef-checkthat:2022:task1:Bilel_Taboubi} \\
Attention-worthy   & CT–CWT–22 & F1$_{Weighted}$ & 0.125 & 0.148   & 0.258 & 0.257  & \textbf{0.412}  & 0.206 \cite{clef-checkthat:2022:task1}  \\
\midrule
\multicolumn{9}{c}{\textbf{Semantics}} \\ \midrule
STS & STS2017-Track 1   & PC  & 0.005 & 0.537   & 0.799 & \textbf{0.813} & 0.809 & 0.754 \cite{cer-etal-2017-semeval}  \\
STS & STS2017-Track 2   & PC  & -0.136 &  0.512   & 0.828 & 0.848  & \textbf{0.857}  & 0.749 \cite{cer-etal-2017-semeval}  \\
STS QS (Q2Q) & Mawdoo3 Q2Q  & F1$_{Micro}$   & 0.491 & 0.910   & 0.816 & 0.895  & 0.935 & \textbf{0.959} \cite{seelawi2019nsurl}  \\
XNLI (Arabic)  & XNLI & Acc & 0.332 & 0.500   & 0.489 & 0.753  & \textbf{0.774}  & 0.713 \cite{artetxe2019cross}    \\
\midrule
\multicolumn{9}{c}{\textbf{Question answering (QA)}} \\\midrule
QA  & ARCD & F1$_{(EM)}$   & 0.085 & 0.368   & 0.502 & \textbf{0.705} & 0.704 & 0.613 \cite{mozannar2019neural}  \\
QA  & MLQA & F1$_{(EM)}$   & 0.066 & 0.377   & 0.376 & 0.620  & \textbf{0.653}  & 0.548 \cite{lewis2019mlqa}   \\
QA  & TyDi QA  & F1$_{(EM)}$   & 0.111 & 0.456   & 0.480 & 0.744  & 0.739 & \textbf{0.820} \cite{tydiqa} \\
QA  & XQuAD & F1$_{(EM)}$   & 0.047 & 0.367   & 0.442 & \textbf{0.729} & 0.722 & 0.665 \cite{artetxe2019cross} \\  
\bottomrule
\end{tabular}%
}
\vspace{-0.2cm}
\caption{Results on NLP tasks. QS: Question similarity, PC: Pearson Correlation, 
JS: Jaccard Similarity, EM: Exact match, POS: positive class. Best result per row is \textbf{boldfaced}. 
\textbf{NA}: experiments could not be performed due to a lack of training data. 
BLOOMZ does not understand some tasks at all as marked with $\ddag$ symbol. \(\varnothing\) - no SOTA results. For the semantic similarity tasks, the negative (``-'') results with random baseline indicate the value of the Pearson correlation, which is between -1 to 1.  
}
\vspace{-0.4cm}
\label{tab:exp_results_nlp_tasks}
\end{table*}

\begin{table}[!ht]
\centering
\setlength{\tabcolsep}{2pt}
\resizebox{0.9\columnwidth}{!}{%
\begin{tabular}{@{}llrrrrrr@{}}
\toprule
\multicolumn{1}{c}{\textbf{Dataset}} & \multicolumn{1}{c}{\textbf{Dialect}} & \multicolumn{1}{c}{\textbf{\#Sent.}} & \multicolumn{1}{c}{\textbf{BLOOMZ}} & \multicolumn{1}{c}{\textbf{Jais}} & \multicolumn{1}{c}{\textbf{\begin{tabular}[c]{@{}c@{}}GPT-3.5\end{tabular}}} & \multicolumn{1}{c}{\textbf{\begin{tabular}[c]{@{}c@{}}GPT-4\end{tabular}}} & \multicolumn{1}{c}{\textbf{SOTA}} \\ \midrule
APT & LEV & 1000 & 11.38 & 13.13 & 18.55 & 17.77 & \textbf{21.90} \\
APT & Nile & 1000 & 12.95 & 16.31 & 21.58 & 18.99 & \textbf{22.60} \\
MADAR & Gulf & 16000 & 32.34 & 34.44 & 34.60 & \textbf{36.18} & 32.46 \\
MADAR & LEV & 12000 & 31.36 & 33.30 & 33.42 & \textbf{35.24} & 32.45 \\
MADAR & MGR & 14000 & 23.59 & 27.61 & 23.91 & \textbf{27.83} & 23.14 \\
MADAR & MSA & 2000 & 42.33 & 38.54 & 37.55 & 37.67 & \textbf{43.40} \\
MADAR & Nile & 8000 & 34.87 & 36.50 & 36.97 & \textbf{37.93} & 35.15 \\
MDC & LEV & 3000 & 10.00 & 14.22 & 17.38 & 16.05 & \textbf{17.63} \\
MDC & MGR & 1000 & 8.28 & 12.80 & 14.46 & \textbf{14.20} & 13.90 \\
MDC & MSA & 1000 & 15.75 & 17.45 & 21.05 & 19.34 & \textbf{20.40} \\
Media & Gulf & 467 & 14.22 & 17.18 & 22.68 & \textbf{22.76} & 19.60 \\
Media & LEV & 250 & 7.54 & 14.94 & 17.65 & 16.65 & \textbf{16.80} \\
Media & MGR & 526 & 4.87 & 11.05 & \textbf{11.58} & 10.20 & 9.60 \\
Media & MSA & 1258 & 20.66 & 28.59 & 35.34 & \textbf{33.57} & 32.65 \\
Bible & MGR & 1200 & 17.09 & 20.96 & 16.72 & 15.29 & \textbf{29.00} \\
Bible & MSA & 1200 & 22.91 & 24.17 & 22.08 & 17.53 & \textbf{31.20} \\ \midrule
\textbf{Avg} & & &\textbf{19.38} &	\textbf{24.09}	& \textbf{23.57} & \textbf{22.57} & \textbf{25.12} \\\bottomrule
\end{tabular}
}
\vspace{-0.2cm}
\caption{BLEU score on MT 
using zero-shot prompts. 
\#Sent: number of test set sentences. 
SOTA results are reported in \citet{sajjad2020arabench}. 
}
\vspace{-0.3cm}
\label{tab:exp_results_mt_only}
\end{table}


\paragraph{Word Segmentation, Syntax and Information Extraction.}
As Table~\ref{tab:exp_results_nlp_tasks} shows, for almost all tasks in this group, the performance is significantly below SOTA performance. For example, the difference between SOTA and GPT-4 (zero-shot) ranges from 6.3\% (segmentation) to 57.6\% (lemmatization). 

\paragraph{Machine Translation.} Table \ref{tab:exp_results_mt_only} reports MT results 
by averaging them dialect-wise for different datasets. Appendix~\ref{app:results:mt} reports detailed results.
The results indicate the short-coming of LLMs when explored with standard and dialectal Arabic. 
\paragraph{Sentiment, Stylistic and Emotion Analysis.} 
In the second group of Table \ref{tab:exp_results_nlp_tasks}, we report results for sentiment, emotion, stance and sarcasm detection mainly over tweets. We observe that on average, performance gap significantly reduced between GPT-4 (best of zero- and few-shot) vs. SOTA compared to GPT-3.5 vs. SOTA, 8.28\% vs 16.44\%, respectively. For sarcasm detection task with ArSarcasm dataset, GPT-4 even outperformed SOTA by 4.41\%. 


\paragraph{News Categorization.} 
Table \ref{tab:exp_results_nlp_tasks} shows that performance gap reduced significantly ranging from 7.1\% to 5.3\% for GPT-3.5 to GPT-4, respectively. Low performance on tweet dataset (ASND) might be due to the higher number of class labels. 

\paragraph{Demographic/Protected Attributes.}
Among the three tasks in this group, two (namely ``name info'' and ``location'' identification) demonstrate a significant performance improvement of over 4.7\% compared to the state-of-the-art (SOTA) results, using the GPT-4 model. 



\paragraph{Ethics and NLP: Factuality, Disinformation and Harmful Content Detection.} 

Across eleven tasks, the performance gap significantly reduced with GPT-4 model, however in some tasks, model's performance is significantly lower than the SOTA. For example, for factuality with COVID-19 disinformation dataset, GPT-4 model's performance is 34\% lower than the SOTA, even though performances of GPT-4 significantly improved compared to GPT-3.5. This task is generally challenging requiring deep contextual analysis and reasoning abilities, and domain knowledge in many of the cases. With a few demonstrations (3-shots) may not be enough to determine the factuality of the content.

\begin{table}[t]
\centering
\scalebox{0.7}{
\begin{tabular}{lcccc}
\toprule
\begin{tabular}[l]{@{}c@{}} \textbf{Dataset} \\ \textit{dom./dial.} \end{tabular}& \textbf{Models} & \textbf{Zero-Shot} & \begin{tabular}[c]{@{}c@{}}\textbf{N-Shot} \\ (\textit{2hrs})\end{tabular} & \textbf{SOTA} \\ \midrule\midrule
\multirow{4}{*}{\begin{tabular}[c]{@{}l@{}}MGB2 \\ {\small \textit{Broadcast/MSA}}\end{tabular}} & W.S & 46.70 & 36.8 & \multirow{4}{*}{\begin{tabular}[c]{@{}c@{}}O: \textbf{11.4} \\ S:11.9\end{tabular}} \\
 & W.M & 33.00 & - &  \\
 & W.Lv2 & 26.20 & 18.8 &  \\
 & USM & 15.70 & \textit{N/A} &  \\ \midrule\midrule
\multirow{4}{*}{\begin{tabular}[c]{@{}l@{}}MGB3 \\ {\small \textit{Broadcast/EGY}}\end{tabular}} & W.S & 83.20 & 77.5 & \multirow{4}{*}{\begin{tabular}[c]{@{}c@{}}O: \textbf{21.4} \\ S: 26.70\end{tabular}} \\
 & W.M & 65.90 & - &  \\
 & W.Lv2 & 55.60 & 44.6 &  \\
 & USM & 22.10 & \textit{N/A} &  \\ \midrule\midrule
\multirow{4}{*}{\begin{tabular}[c]{@{}l@{}}MGB5 \\ {\small \textit{Broadcast/MOR}}\end{tabular}} & W.S & 135.20 & 114.6 & \multirow{4}{*}{\begin{tabular}[c]{@{}c@{}}O: \textbf{44.1}\\ S:49.20\end{tabular}} \\
 & W.M & 116.90 & - &  \\
 & W.Lv2 & 89.40 & 85.5 &  \\
 & USM & 51.20 & \textit{N/A} &  \\ \midrule\midrule
\multirow{4}{*}{\begin{tabular}[c]{@{}l@{}}QASR.CS \\ {\small \textit{Broadcast/Mixed}}\end{tabular}} & W.S & 63.60 & - & \multirow{4}{*}{\begin{tabular}[c]{@{}c@{}}O: \textbf{23.4} \\ S: 24.90\end{tabular}} \\
 & W.M & 48.90 & - &  \\
 & W.Lv2 & 37.90 & 31.2$^+$ &  \\
 & USM & 27.80 & \textit{N/A} &  \\ \midrule\midrule
\multirow{4}{*}{\begin{tabular}[c]{@{}l@{}}DACS \\ {\small \textit{Broadcast}}\\/{\small \textit{MSA-EGY}}\end{tabular}} & W.S & 61.90 & - & \multirow{4}{*}{\begin{tabular}[c]{@{}c@{}}O: 15.9 \\ S: 21.3\end{tabular}} \\
 & W.M & 48.70 & - &  \\
 & W.Lv2 & 34.20 & 30.4$^+$ &  \\
 & USM & \textbf{14.30} & \textit{N/A} &  \\ \midrule\midrule
\multirow{4}{*}{\begin{tabular}[c]{@{}l@{}}ESCWA.CS \\ {\small \textit{Meeting/Mixed}}\end{tabular}} & W.S & 101.50 & - & \multirow{4}{*}{\begin{tabular}[c]{@{}c@{}}O: 49.8 \\ S:48.00\end{tabular}} \\
 & W.M & 69.30 & - &  \\
 & W.Lv2 & 60.00 & 53.6$^+$ &  \\
 & USM & \textbf{45.70} & \textit{N/A} &  \\ \midrule\midrule
\multirow{4}{*}{\begin{tabular}[c]{@{}l@{}}CallHome \\ {\small \textit{Telephony/EGY}}\end{tabular}} & W.S & 155.90 & 152.9 & \multirow{4}{*}{\begin{tabular}[c]{@{}c@{}}O: \textbf{45.8}* \\ S: 50.90\end{tabular}} \\
 & W.M & 113.70 & - &  \\
 & W.Lv2 & 78.70 & 64.6 &  \\
 & USM & 54.20 & \textit{N/A} & \\\bottomrule
\end{tabular}
}
\vspace{-0.2cm}
\caption{Reported WER ($\downarrow$) on ASR 
in zero and few-shot setup and domain-specific ASR setup. W.{S,M,Lv2} stands for OpenAI Whisper small, medium and Largev2 model. O: represent offline; S: streaming ASR; * represent the model's input is 8kHz sampling rate and Offline model was re-trained to accommodate telephony data. $^+$ represent model fine-tuned with 2hrs of MGB2-data.}
\label{tab:exp_results_asr_only}
\vspace{-0.25cm}
\end{table}

\begin{table}[!h]
\centering
\scalebox{0.8}{
\begin{tabular}{l|ccc|cc}
\toprule
\multicolumn{1}{c}{} & \multicolumn{3}{|c|}{\textbf{Subjective (MOS)} $\uparrow$} & \multicolumn{2}{c}{\textbf{Objective} $\downarrow$} \\ 
\hline 
\multicolumn{1}{c|}{Model} & Diac. & Natur. & Intel. & WER & CER \\ \hline
Amazon & 8.2 & 8.3 &  9.8 &  5.2 & \bf 1.0 \\
Q-K & \bf 9.5 & 8.6 &  9.8 & \bf 3.7 & 1.2 \\
\bottomrule
\end{tabular}}
\vspace{-0.2cm}
\caption{
Evaluation for Arabic TTS. Diac.: Diacritization, Natur.: Naturalness, Intel.: Intelligibility.}
\label{tab:exp_results_tts_only}
\vspace{-0.4cm}
\end{table}

\paragraph{Semantics:}

The results for various semantic tasks reported in Table \ref{tab:exp_results_nlp_tasks} indicate that the performance on three out of the four tasks surpasses the SOTA, with an overall improvement of 9.9\%.



\paragraph{Question answering (QA):} 
Results on four QA datasets (Table~\ref{tab:exp_results_nlp_tasks}) show that for three of them, GPT-4 achieved higher performance than SOTA with an overall improvement of 9.2\%. 

\paragraph{Performance of the Jais Model:} 
In Table \ref{tab:exp_results_mt_only} (MT only) and Table \ref{tab:performance_jais} (see Appendix~\ref{ssec:app_performance_jais}), we report the performance of the Jais model alongside a comparison with other models in a zero-shot setting. The results presented in the table indicate that, on average, the performance of the Jais model outperform that of both random and BLOOMZ models. However, it underperforms compared to the models developed by OpenAI. For the QA task, the Jais model's performance is 4\% better than that of GPT-3.5. It is surprising that the performance of the news categorization task is significantly lower with the Jais model. The reason for this is that the model most often incorrectly predicts texts about politics as belonging to ``crime, war, and conflict''.  
%

\subsection{Speech Recognition and Synthesis}
\label{ssec:results_asr}


In Table \ref{tab:exp_results_asr_only}, we reported the performance of ASR using different datasets and models. We observed that USM outperforms Whisper in all datasets in both zero and few-shot settings. The USM model performs comparably to standard task- and domain-specific ASR systems and is better equipped to handle cross-language and dialectal code-switching data from unseen domains compared to the SOTAs and Whispers few-shot fine-tuned model. 

Both the subjective and objective evaluations for the TTS are reported Table \ref{tab:exp_results_tts_only}.  
The results show that Q-K model \cite{abdelali-2022-natiq} outperformed Amazon Polly significantly in objective evaluation (WER). Subjective scores show Q-K is better in naturalness and diacritization. With almost similar performance in intelligibility.






\section{Findings}
\label{ssec:discussion}



\paragraph{NLP Model Performances.} 




Our qualitative analysis revealed certain patterns of errors in sequence tagging tasks like segmentation, POS tagging, and NER. These patterns encompassed: i) deviations in the output format, ii) instances where responses included extra or omitted tokens, and iii) cases where the model generated output labels in Arabic instead of English. Notably, these errors occasionally led to a noticable drop in the performance of LLMs. In certain multilabel tasks, such as propaganda detection, the models occasionally produced outputs that fell outside the predefined set of labels. This findings suggests that LLMs may not be seamlessly deployable, demanding considerable effort in crafting prompts to attain precise outputs or engaging in post-processing to align outputs with reference labels. In essence, these findings highlight the intricate nature of utilizing LLMs in sequence tagging tasks, emphasizing the need for a careful handling and optimization in real-world applications.

Our comprehensive study highlights the disparities in performance of LLMs -- GPT-3.5 and GPT-4, as compared to SOTA models, in zero and few-shot settings. 
GPT-3.5 exhibits a significant performance gap when compared to SOTA. However, GPT-4 manages to narrow this gap to some extent and even outperforms the SOTA models in high-level abstract tasks such as STS, QA, claim detection, news categorization, demographic attributes, and XNLI.
Moreover, GPT-4 outperforms GPT-3.5 across all tasks.
However, it remains a challenge for GPT-4 to surpass SOTA performance consistently in sequence tagging (especially syntactic and segmentation) tasks.
The performance of BLOOMZ is significantly lower than SOTA and GPT models, and in some cases lower than random baseline. 
The performances of both open and close models are heavily dependent on the \textit{effective prompt} and implementing appropriate \textit{post-processing techniques}.
Overall, these findings indicate the potential of GPT-4 as a \textit{multi-task model} without heavily relying on task-specific resources, particularly in zero/few-shot settings.

The \textit{few-shot results} across \textit{seven different datasets} show an average improvement from 0.656 (0-shot) to 0.721 (10-shot) indicating the promise of few-shot learning, as depicted in Figure \ref{fig:few_shot_extended_exp} (in Appendix), with individual results are reported in Table \ref{tab:few_shot_extended}.

\begin{table}[ht]
\centering
\setlength{\tabcolsep}{3pt}
\resizebox{\columnwidth}{!}{%
\begin{tabular}{@{}lllrrrr@{}}
\toprule
\multicolumn{1}{c}{\textbf{Task Name}} & \multicolumn{1}{c}{\textbf{Dataset}} &\multicolumn{1}{c}{\textbf{Metric}} & \multicolumn{1}{c}{\textbf{\begin{tabular}[c]{@{}c@{}}0-shot \\ \end{tabular}}} & \multicolumn{1}{c}{\textbf{\begin{tabular}[c]{@{}c@{}}3-shot\end{tabular}}} & \multicolumn{1}{c}{\textbf{\begin{tabular}[c]{@{}c@{}}5-shot\end{tabular}}} & \multicolumn{1}{c}{\textbf{\begin{tabular}[c]{@{}c@{}}10-shot\end{tabular}}} \\ \midrule
NER & ANERcorp & M-F1 & 0.355 & 0.420 & 0.426 & 0.451 \\
Sentiment & ArSAS & M-F1 & 0.569 & 0.598 & 0.619 & 0.639 \\
News Cat. & ASND & M-F1 & 0.667 & 0.594 & 0.674 & 0.723 \\
Gender & Arap-Tweet & M-F1 & 0.868 & 0.980 & 0.931 & 0.937 \\
Subjectivity & In-house & M-F1 & 0.677 & 0.745 & 0.740 & 0.771 \\
XNLI (Ar) & XNLI & Acc & 0.753 & 0.774 & 0.789 & 0.809 \\
QA & ARCD& F1/EM & 0.705 & 0.704 & 0.718 & 0.716 \\ \midrule
\textbf{Average} & & \ & \textbf{0.656} & \textbf{0.688} & \textbf{0.700} & \textbf{0.721} \\ \bottomrule
\end{tabular}%
}
\vspace{-0.2cm}
\caption{Results from few-shot experiments over seven tasks with GPT-4. M-F1: Macro-F1, Ar: Arabic, EM: exact match}
\label{tab:few_shot_extended}
\vspace{-0.3cm}
\end{table}




The use of \textit{native language prompts} with GPT-4 in a zero-shot context highlighted the role played by the prompt language, as we observed increased performance (1\%) in three out of seven datasets compared to their counterparts with English prompts while  two underperformed, and one showed equivalent performance (see Table \ref{tab:prompt_native_english} in Appendix).

When evaluating these LLMs in \textit{multi-dialectal} settings, the performance gap between MSA and dialectal test sets becomes more evident.
For example, in both the GPT-models, we noticed a large discrepancy in the 
POS accuracy of 0.367 \textit{vs.} 0.323 on MSA and dialects respectively. 
Similarly, for the dialect identification we notice a significant difference between the SOTA acoustic and lexical model with respect to LLMs results.

From the average \textit{performance gap between semantic and syntactic tasks}, as reported in Table \ref{tab:semantic_syntactic_task},
we noticed the discrepancy in semantic tasks is much lower than in syntactic tasks, across the three LLMs. This suggests that these models might be better equipped at encoding and expressing semantic information than in pinpointing specific syntactic phenomena in their inputs.
Moreover, these performance gaps can also be linked to 
\textit{undesirable hallucination}. In particular, during the MT for the Bible, results reveal an interesting phenomenon. It appears that the GPT models, particularly GPT-3.5-turbo, tend to hallucinate and insert additional content in their responses. 

\textit{Is the data contaminated?}
We have used some datasets for evaluation that are released after the cut-off date of ChatGPT training (September 2021), which include subjectivity, propaganda, check worthiness, factuality (CT-CWT-22), harmful content, and attention worthiness. Moreover, we experiment with nine datasets using the guided instructions approach proposed by \citet{golchin2023time} revealing that GPT-4 could not produce any example from these datasets. Thus, 
we can confirm that the models have not been contaminated with such datasets as detailed in 
in Appendix~\ref{ssec:appx_data_contamination}.

\paragraph{Speech Model Performances:} 
We observed the performance of these models is heavily dependent on the architecture parameters. USM model performs comparably with SOTA for MSA. Both Whisper (and its variants) and USM show a performance gap when dealing with dialects, specially Moroccan dialect. Fine-tuning the open model (Whisper Largev2) with only 2 hours of speech data bridges the performance gap significantly, indicating the potential to be a robust and strong foundation model. Our observation also suggests that USM model is better equipped to handle code-switching phenomena in spoken utterance than the supervised large transformer models.

\section{Related Work}
\label{sec:related_work}


\paragraph{Models for NLP:}
Since the inception of the transformer architecture \cite{vaswani2017attention}, there have been efforts to develop larger models with its variants such as BERT~\cite{devlin2019bert}, RoBERTa~\cite{liu2019roberta}, XLM-RoBERTa~\cite{conneau2019unsupervised}, GPT models \cite{Radford2018ImprovingLU,radford2019language,ouyang2022training} among others. 

Such advancements have led to the development LLMs with parameter sizes exceeding 100 billion, which are pre-trained on massive datasets. Examples of LLMs include Megatron~\cite{shoeybi2019megatron}, GPT-3~\cite{brown2020language}, GPT-Jurassic~\cite{lieber2021jurassic}, OPT-175B \cite{zhang2022opt}, and Bloom~\cite{scao2022bloom}. This unprecedented scale enabled new capabilities that address the zero-shot and multilingual tasks learning. ChatGPT (GPT-3.5) and its subsequent model GPT-4 is the latest development in NLP that have addressed many limitations of prior LLMs and enabled us to perform diverse tasks \cite{openai2023gpt4}. The ability of LLMs to solve various tasks can be attributed to the meticulous design of prompts, which enable the generation of desired responses~\cite{weichain,shin2020autoprompt}.

\noindent
\textbf{Models for Speech Processing:}
When handling complex audio/speech data, LLMs face significant challenges. However, with the advent of self-supervised learning, models like Wav2vec, WavLM, and Whisper have been leading in addressing these challenges ~\cite{baevski2019vq, baevski2020wav2vec, chen2022wavlm, radford2022robust}. More recent developments like the USM and VALL-E have demonstrated superior capabilities in ASR and zero-shot TTS tasks, respectively ~\cite{zhang2023google, wang2023neural}.
\noindent
\textbf{LLMs Benchmarking:}
Since the release of ChatGPT, there have been efforts to evaluate the performance of LLMs on standard NLP tasks \cite{bubeck2023sparks,bang2023multitask,ahuja2023mega,hendy2023good}. 
\citet{liang2022holistic} conducted a comprehensive assessment of LLMs for English. It encompassed various metrics such as accuracy, calibration, toxicity, and efficiency, along with 42 scenarios involving 30 prominent language models. 

\noindent
\textbf{Benchmarks on Arabic:}
The complexity and linguistic diversity of Arabic have led to a limited number of benchmarks for language tasks, such as ORCA \cite{elmadany2022orca}, ALUE \cite{seelawi2021alue}, ArBERT \cite{abdul2021arbert}, and AraBench \cite{sajjad2020arabench}.

\noindent
\textbf{LAraBench:} To the best of our knowledge, our study represents the first comprehensive Arabic language benchmarking effort exploring GPT-3.5 (zero-shot), GPT-4 (zero- and few-shot), BLOOMZ 
 (zero-shot), Jais (zero-shot) and Speech models like Whisper and USM. Our evaluation spans a broad array of LLMs, tasks, and datasets, distinguishing it from prior benchmarks in terms of task and dataset diversity, test setup, modalities (text, speech), and state-of-the-art comparisons. Table \ref{tab:prev_studies_comparison} (Appendix \ref{sec:appx_benchmark_on_arabic}), provides a detailed comparison.

\section{Conclusion and Future Studies}
\label{sec:conclusion}

This study is the \textit{first} large-scale benchmark that brings together both 
Speech and NLP tasks under the same study. We report the performance of LLMs 
covering different domains and dialects. Our study also considers tasks with a wide range of complexity ranging from token to text classification, different application settings, NER to sentiment, factuality and disinformation, ASR, and TTS among others. We evaluate 33 tasks and 61 datasets with 98 test setups, which are very prominent for Arabic AI. We compare and report the performance of each task and dataset with SOTA, which will enable the community and practitioners of large language models to decide on their uses of these models. 
Future work aims to investigate open models and explore ways to reduce the performance gap with SOTA; enhance prompts for better performance; and expand datasets and tasks.




\section*{Limitations}


The main focus of this study was to benchmark LLMs for Arabic NLP and Speech tasks. 
We evaluated several large models, including those from OpenAI, BLOOMZ, Jais, USM, and Whisper, and compared them to the SOTA. We plan to extend our study by adding other models recently released for Arabic.
In this work, we benchmarked 61 datasets with 98 test setups for 33 tasks. However, we did not benchmark all available data sets. For example, the study reported in \cite{elmadany2022orca} benchmarked 19 sentiment datasets, whereas we only covered one. It is also possible that we missed many other Arabic NLP and Speech tasks, which we will attempt to cover in the future. Our current results are highly dependent on prompt design. Additional efforts on prompt engineering could potentially improve the results. 

In addition, performance may vary depending on the version of the models we used.\footnote{\url{https://platform.openai.com/docs/models/overview}} For GPTs, we utilized gpt-3.5-turbo-0301 and gpt-4-0314 versions for our NLP tasks. To ensure transparency and reproducibility, we made all resources publicly available. 
This will facilitate the easy replication of our results using the provided pipeline and the fixed model versions. The same principle extends to our speech models. We have taken steps to maintain versioning not only for the models themselves but also for the prompts used. This ensures that our work remains reproducible for future researchers in the field.

\paragraph{Potential Risk}
We do not oversee any potential risk that can result from our study. 

\section*{Ethics Statement}
\label{sec:ethics_statement}
Our evaluation includes tasks and datasets related to disinformation, and hate speech. We used publicly available datasets and evaluated whether LLMs can classify them (e.g., hate vs. non-hate). We do not foresee any potential risk from the outcome of our work. 

\section*{Acknowledgments}
The contributions of M. Hasanain were funded by the NPRP grant 14C-0916-210015, which is provided by the Qatar National Research Fund (a member of Qatar Foundation).

\bibliography{bib/anthology,bib/bibliography}

\appendix

\section*{Appendix}
\label{sec:appendix}
\appendix
\section{Tasks and Datasets}
\label{sec:appx-tasks-datasets}

In this section, we discuss the tasks and the associated datasets by grouping them based on ACL-2022 track.\footnote{\url{https://www.2022.aclweb.org/callpapers}} In Tables ~\ref{tab:datasets-txt} and ~\ref{tab:datasets-speech}, we provide a summarized description of the test sets used for evaluating textual and speech processing tasks, respectively.

\begin{table*}[] 
\centering
\setlength{\tabcolsep}{6pt}
\resizebox{0.7\textwidth}{!}{
\begin{tabular}{llll}
\toprule
\textbf{Dataset} & \textbf{Task} &\textbf{Domain} & \textbf{Test Set Size} \\ \midrule
\multicolumn{4}{c}{\textbf{Word Segmentation, Syntax and Information Extraction}} \\
\midrule
WikiNews	&	Segmentation	&	News articles (MSA)	&	400 sentences \\
\citet{samih2017learning}	&	Segmentation	&	\multicolumn{1}{p{4cm}}{Tweets (Dialects: EGY, LEV, GLF, MGR)}	&	70 X 4 dialects \\
WikiNews	&	Lemmatization	&	News articles (MSA)	&	400 sentences \\
WikiNews	&	Diacritization	&	News articles (MSA)	&	400 sentences \\
\citet{darwish2018diacritization}	&	Diacritization	&	\multicolumn{1}{p{4cm}}{Sentences (Dialects: Moroccan, Tunisian)}	&	1,640 X 2 dialects\\
WikiNews	&	POS	&	News articles (MSA)	&	400 sentences \\
\citet{samih2017learning}	&	POS	&	\multicolumn{1}{p{4cm}}{Tweets (Dialects: EGY, LEV, GLF, MGR)}	&	70 X 4 dialects \\
XGLUE (Arabic)	&	POS	&	Web, Wikipedia	&	680 sentences\\
Conll2006	&	Parsing	&	MSA	&	146 sentences \\
ANERcorp	&	NER	&	News articles	&	924 sentences \\
AQMAR	&	NER	&	Wikipidia	&	1,976 sentences \\
QASR	&	NER	&	Transcripts	&	7,906 segments \\
QADI	&	Dialect	&	Tweets	&	3,797 \\
ADI	&	Dialect	& \multicolumn{1}{p{4cm}} {\small Transcripts (Dialects: EGY, IRA, JOR, KSA, KUW, LEB, LIB, MOR, PAL, QAT, SUD, SYR, UAE, YEM, and MSA)}	&	750 \\
\midrule
\multicolumn{4}{c}{\textbf{Sentiment, Stylistic and Emotion Analysis}} \\
\midrule
ArSAS	&	Sentiment	&	Tweets	&	4,213 \\
SemEval2018-Task1	&	Emotion	&	Tweets (Dialectal)	&	1,518 \\
Unified-FC	&	Stance	&	News articles	&	3,042 claim-article pairs\\
ANS	&	Stance	&	News articles	&	379 headline pairs \\
ArSarcasm	&	Sarcasm	&	Tweets	&	2,110 \\
ArSarcasm-2	&	Sarcasm	&	Tweets	&	3,000 \\
\midrule	
\multicolumn{4}{c}{\textbf{News Categorization}} \\\midrule
ASND	&	News Cat.	&	Posts$*$	&	1,103\\
SANAD/Akhbarona	&	News Cat.	&	News articles	&	7,843 \\
SANAD/AlArabiya	&	News Cat.	&	News articles	&	7,125 \\
SANAD/AlKhaleej	&	News Cat.	&	News articles	&	4,550 \\
\midrule	
\multicolumn{4}{c}{\textbf{Demographic Attributes}} \\\midrule
ASAD	&	Name Info	&	Wikidata	&	80,130 \\
UL2C	&	Location	&	User loc. (Twitter)	&	28,317 \\
Arap-Tweet	&	Gender	&	Usernames (Twitter)	&	640 \\
\midrule	
\multicolumn{4}{c}{\textbf{Ethics in NLP: Factuality, Disinformation and Harmful Content Detection}} \\ \midrule	
OffensEval2020	&	Offensive lang.	&	Tweets (Dialectal)	&	2,000 \\
OSACT2020	&	Hate Speech	&	Tweets (Dialectal)	&	2,000 \\
ASAD	&	Adult Content	&	Tweets (Dialectal)	&	10,000 \\
ASAD	&	Spam	&	Tweets (Dialectal)	&	28,383 \\
In-house	&	Subjectivity	&	News articles	&	297 sentences \\
WANLP23	&	Propaganda	&	Tweets	&	323 \\
CT–CWT–22	&	Checkworthiness	&	Tweets (COVID19)	&	680 \\
COVID19 Disinfo.	&	Factuality	&	Tweets	&	996 \\
Unified-FC	&	Factuality	&	News articles	&	422 claims \\
ANS	&	Factuality	&	News articles	&	456 headlines \\
CT–CWT–22	&	Claim	&	Tweets (COVID19)	&	1,248 \\
CT–CWT–22	&	Harmful content	&	Tweets (COVID19)	&	1,201 \\
CT–CWT–22	&	Attention-worthy	&	Tweets (COVID19)	&	1,186 \\
\midrule	
\multicolumn{4}{c}{\textbf{Semantic Textual Similarity (STS)}} \\\midrule
STS2017-Track 1	&	STS	&	Image captions	&	250  sentence pairs \\
STS2017-Track 2	&	STS	&	Image captions	&	250  sentence pairs\\
Mawdoo3 Q2Q	&	\begin{tabular}[c]{@{}l@{}}STS QS (Q2Q)\end{tabular}	&	Questions	&	3,715 question pairs \\
XNLI	&	XNLI	&	ANC	&	5,010  sentence pairs \\
\midrule	
\multicolumn{4}{c}{\textbf{Question Answering (QA)}} \\\midrule
ARCD	&	QA	&	Wikipedia	&	702 questions\\
MLQA	&	QA	&	Wikipedia	&	5,335 questions\\
TyDi QA	&	QA	&	Wikipedia	&	921 questions\\
XQuAD	&	QA	&	Wikipedia	&	1,190 questions\\
\bottomrule
\end{tabular}
}
\caption{Summary on test sets and their sizes used in evaluation for the different textual tasks. \textbf{ANC}: American National Corpus. \textbf{Posts$*$}: posts from Twitter, Youtube and Facebook. \textbf{News Cat.}: News Categorization}
\label{tab:datasets-txt}
\end{table*}

\subsection{Word Segmentation, Syntax and Information Extraction}
\subsubsection{Segmentation}    
Segmentation is an important problem for language like Arabic, which is rich with bound morphemes that change the tense of verbs, or represent pronouns and prepositions in nouns. It is a building block for NLP tasks such as search, part-of-speech tagging, parsing, and machine translation. The idea is segmenting Arabic words into  prefixes, stems, and suffixes, which can facilitate 
many other tasks.

\subsubsection*{Datasets}
\paragraph{WikiNews} For modern standard Arabic (MSA), we used the WikiNews dataset of~\citep{darwish2016farasa} which comprises 70 news articles in politics, economics, health, science and technology, sports, arts, and culture. The dataset has 400 sentences (18,271 words) in total. 

\paragraph{Tweets} For the dialectal Arabic, we used the dataset reported in~\citep{samih2017learning}, which provides 1400 tweets in Egyptian, Gulf, Levantine, and Maghrebi dialects for a total of 25,708 annotated words \label{wikinews}.


        
\subsubsection{Part-Of-Speech (POS) Tagging} 
Part-of-speech (POS) is one of the fundamental components in the NLP pipeline. It helps in extracting higher-level information such as named entities, discourse, and syntactic parsing.

\subsubsection*{Datasets}
\paragraph{WikiNews} We used for this task the WikiNews dataset tagged for POS~\citep{darwish2017arabic} for modern standard Arabic.
\paragraph{Tweets} For POS tagging with noisy texts and different dialects we used the same dataset reported in~\citep{samih2017learning} (see \S\ref{wikinews}).

\paragraph{XGLUE} We also used the Arabic part of XGLUE benchmark~\cite{liang2020xglue} for POS tagging, which uses a subset of Universal Dependencies Treebanks (v2.5)~\citep{zeman2020universal}.
    
\subsubsection{Lemmatization} 
Lemmatization is another component in the NLP pipeline, which reduces words to their base or root form, known as a lemma. It takes into consideration the morphological analysis of the words, which uses the context and POS to convert a word to its simplest form.
This task differs from segmentation which only separates a word stem from prefixes and suffixes. In contrast, lemmatization requires returning the lexicon entry for a certain word, which may depend on POS tagging.

\paragraph{Dataset}
We used WikiNews dataset tagged for lemmas ~\citep{mubarak-2018-build} (see \S\ref{wikinews} for the details of the dataset).

        
\subsubsection{Diacritization} 
\label{ssec:diacritization}
Diacritization involves assigning the diacritics to each letter in an Arabic word within a sentence. Diacritical marks indicate the correct pronunciation and meaning of the written Arabic words. For example, different word diacretizations could transform a noun into a verb or vice versa.

\subsubsection*{Datasets}
\paragraph{WikiNews} We use a dataset of Modern Standard Arabic from~\citep{mubarak2019highly} that comprises  fully diacritized WikiNews corpus~\cite{darwish-etal-2017-arabic}.
\paragraph{Bibles} This dataset includes translations of the New Testament into two Maghrebi sub-dialects: Moroccan and Tunisian \cite{darwish2018diacritization,abdelali2019diacritization}. 

\subsubsection{Parsing} Dependency parsing is the task of identifying syntactical and grammatical relations among the words in a sentence. These dependencies result in a hierarchical tree representation that captures the structure of the sentence at different levels. 

\paragraph{Dataset} For this task we used the Arabic part of CoNLL-X 2006 shared tasks on dependency parsing~\citep{buchholz2006conll}, which has 4,990 scoring tokens and uses the Prague Arabic Dependency Treebank~\citep{hajic2004prague}.
        
        
\subsubsection{Named-Entity Recognition (NER)} This task involves identifying and classifying the words in a sentence that are proper names, names of places, entities like organizations or products, amongst other things.  This depends on understanding the context and the relations of a word or a collection of words in a sentence, and is key to tasks such as question answering.  

\subsubsection*{Datasets} 
\paragraph{ANERCorp} We used the test corpus of the ANERCorp dataset~\citep{10.1007/978-3-540-70939-8_13,benajiba2007anersys}, which contains 316 articles, 150,286 tokens and 32,114 types, and classifies words into one of four classes (organization, location, person and miscellaneous), we used the test split of the dataset for our evaluation. 

\paragraph{AQMAR} The dataset is developed as an evaluation suite for the named entity recognition task in Arabic. It consists of a collection of 28 Wikipedia articles with 74,000 tokens.  We consider the articles corresponding to the test split for our evaluation. ~\citep{schneider2012coarse}. 
\paragraph{QASR} 
The QASR dataset consists of 70k words extracted from 2,000 hours of transcribed Arabic speech~\citep{mubarak2021qasr}.

\subsection{Machine Translation (MT)} 
The machine translation evaluation set is a rich set that covers a variety of Arabic in addition to the Modern Standard Arabic (MSA). The genera of the evaluation set also cover formal, informal, speech, and other modalities. These types and varieties allowed us to assess the system and reveal its potential and limitations. For this study, we focused on translating Arabic to English and used the datasets discussed below. 

\subsection*{Datasets}    
\paragraph{MADAR Corpus} This dataset consists of 2,000 sentences from the BTEC corpus translated to modern standard Arabic and four major dialects from 15 countries \citep{bouamor2018madar}.

\paragraph{\citet{zbib2012machine}:} It is collected from the Arabic-Dialect/English Parallel Text (APT), which consists of 2,000 sentences with 3.5 million tokens of translated dialectal Arabic.

\paragraph{Multi-dialectal Parallel Corpus of Arabic (MDC)} This dataset also consists of 2,000 sentences in Egyptian, Palestinian, Syrian, Jordanian, and Tunisian dialects and their English counterparts~\citep{bouamor2014multidialectal}.


\paragraph{The Bible} It consists of 8.2k parallel sentences translated into modern standard Arabic, and to Moroccan\footnote{The Morocco Bible Society https://www.biblesociety.ma} and Tunisian\footnote{The United Bible Societies https://www.bible.com} dialects ~\cite{abdelali2019diacritization}.

\paragraph{Media Dataset}
The dataset consists of 7.5 hours of recordings collected from five public broadcasting channels that cover programs with Maghrebi, Lebanese, Omani dialects, and MSA with genres involving movies, news reports, and cultural programs. The recordings were transcribed and translated by a professional translation house \cite{sajjad2020arabench}.


\subsection{Dialect Identification} 
Dialect is defined as the speaker's grammatical, lexical, and phonological variation in pronunciation \cite{7361147}. Automatic Dialect Identification (ADI) has became an important research area in order to improve certain applications and services, such as ASR and many downstream NLP tasks. 

\paragraph{Dataset} 
For this task, we used the QADI and ADI datasets. QADI consists of a wide range of country-level Arabic dialects covering 18 different countries in the Middle East and North Africa region~\citep{abdelali-etal-2021-qadi}. It consists of 540,590 tweets from 2,525 users. 
The ADI dataset is comprised of 750 utterances obtained from a subset of ADI-5\footnote{\url{https://arabicspeech.org/adi_resources/mgb3}} and ADI-17\footnote{\url{https://arabicspeech.org/adi_resources/mgb5}} test sets. We selected 50 utterances from each of the 14 countries in the Middle East and North Africa region along with MSA utterances. 



\subsection{Sentiment, Stylistic and Emotion Analysis} 

\subsubsection{Sentiment Analysis} 
Sentiment analysis has been an active research area and aims to analyze people’s sentiment or opinion toward entities such as topics, events, individuals, issues, services, products, organizations, and their attributes \cite{liu2012survey,zhang2018deep}. This task involves classifying the content into sentiment labels such as positive, neutral, and negative. 

\paragraph{Dataset}  
ArSAS dataset consists of 21k Arabic tweets covering multiple topics that were collected, prepared, and annotated for six different classes of speech-act labels and four sentiment classes~\cite{elmadany2018arsas}. For the experiments, we used only sentiment labels from this dataset.


\subsubsection{Emotion Recognition} 
Emotion recognition is the task of categorizing different types of content (e.g., text, speech, and visual) in different emotion labels (six basic emotions \cite{ekman1971universals} or more fine-grained categories \cite{demszky-etal-2020-goemotions}).

\paragraph{Dataset}
For the emotion recognition tasks we used SemEval-2018 Task 1: Affect in Tweets \cite{mohammad2018semeval}. The task is defined as classifying a tweet as one or more of the eleven emotion labels, which is annotated as a multilabel (presence/absence of 11 emotions) annotation setting. 

        


\subsubsection{Stance Detection}
Stance is defined as the expression of the speaker’s view and judgment toward a given argument or statement~\cite{biber1988adverbial}. Given that the social media platforms allow users to 
 consume and disseminate information by expressing their views, enabling them to obtain instant feedback and explore others' views, it is important to characterize a stance expressed in a given content.   Automatic stance detection also allows for assessing public opinion on social media, particularly on different social and political issues such as abortion, climate change, and feminism, on which people express supportive or opposing opinions ~\cite{ALDAYEL2021102597,kuccuk2020stance}. The task involves ``classification as the stance of the producer of a piece of text, towards a target as either one of the three classes: \{support, against, neither\} or \{agree, disagree, discuss, or unrelated\}'' \cite{kuccuk2020stance}.

\paragraph{Datasets}

\noindent
\paragraph{Unified-FC} dataset consists of claims collected from Verify.sy (false claims) and Reuters (true claims), which resulted in 422 claims. Based on these claims documents are collected using Google custom search API and filtered by computing claim-documents similarity~\cite{baly2018integrating}. This approach resulted in 3,042 claim-documents pairs, which are then annotated for stance (agree, disagree, discuss, unrelated) by Appen crowd-sourcing platform. 

\noindent
\paragraph{ANS} \citet{khouja-2020-stance} developed a dataset by first sampling news titles from Arabic News Texts (ANT) corpus \cite{chouigui2017ant} and then generating true and false claims. From these claims stance (three classes – agree, disagree, other) is annotated from a pair of sentences using Amazon Mechanical Turk and Upwork. The dataset consists of 3,786 claim-reference pairs.

\noindent
\paragraph{ArSarcasm} \citet{abu-farha-magdy-2020-arabic} developed an Arabic sarcasm detection dataset. The dataset was created using previously available Arabic sentiment analysis datasets ~\cite{rosenthal2017semeval, nabil2015astd} and adds sarcasm and dialect labels to them. The dataset contains 10,547 tweets, 1,682 of which are sarcastic. The training set contains 8,437 tweets, while the test set contains 2,110 tweets.

\noindent
\paragraph{ArSarcasm-v2} This dataset is an extension of the original ArSarcasm dataset published along with the paper ~\cite{farha2020arabic}. ArSarcasm-v2 conisists of ArSarcasm along with portions of DAICT corpus and some new tweets. Each tweet was annotated for sarcasm, sentiment and dialect. The final dataset consists of 15,548 tweets divided into 12,548 training tweets and 3,000 testing tweets. ArSarcasm-v2 was used and released as a part of the shared task on sarcasm detection and sentiment analysis in Arabic.

\subsection{News Categorization} 
News text categorization was a popular task in the earlier days of NLP research \cite{sebastiani2002machine}. The idea of to assign a category $C=\{c_1,...c_n\}$ to a document $D=\{d_1,...d_n\}$. 
For the news categorization the $D$ is a set of news articles and $C$ is a set of predefined categories. Most often a news article can be categorized into more than one category and the models are trained in a multilabel setting. While earlier work mostly focused on news article, however, lately it has been used for the categorization of tweets in which news articles are shared as a part of a tweet. 


\subsubsection*{Datasets} 
\paragraph{Social Media Posts} ASND is a News Tweets dataset ~\citep{chowdhury2020improving}, collected from Aljazeera news channel accounts on Twitter, Facebook, and YouTube. The dataset consists of twelve categories such as art-and-entertainment, business-and-economy, crime-war-conflict, education, environment, health, human-rights-press-freedom, politics, science-and-technology, spiritual, sports, and (xii) others. We used the test split from each dataset for the evaluation.



\paragraph{Arabic News}
SANAD corpus is a large collection of Arabic news articles collected from Akhbarona, AlKhaleej, and AlArabiya~\cite{einea2019sanad}. The dataset has  separate collections gathered from different news media, each of which has six news categories; namely culture, finance, medical, politics, sports and
technology. 






        
\subsection{Demographic/Protected Attributes}
Demographic information (e.g., gender, age, country of origin) are useful in many different applications such as understanding population characteristics, personalized advertising, socio-cultural studies, etc. Demographic information helps governments, businesses, and organizations understand their target audiences, and plan accordingly.

\subsubsection{Gender} 
Gender analysis can reveal important differences between male and female users such as topics of interest, gender gap, preferences, etc.

\paragraph{Dataset}
We used the Arap-Tweet test set, a large-scale and multi-dialectal corpus of tweets from 11 regions and 16 countries in the Arab world, representing the major Arabic dialectal varieties \cite{zaghouani-charfi-2018-arap}.

\subsubsection{Location} 
Identifying user locations is useful for many applications such as author profiling, dialect identification, recommendation systems, etc. Often, users on social media platforms, such as Twitter, declare their locations in noisy ways, and mapping these locations to countries is a challenging task.

\paragraph{Dataset}
We used the UL2C dataset, which contains 28K unique locations, as written by Arabic Twitter users, and their mappings to Arab countries~\cite{mubarak2021ul2c}.

\subsubsection{Name Info} 
Names contain important  information about our identities and demographic characteristics, including factors like gender, nationality, and ethnicity. The purpose of this task is to predict the country of origin of a person name giving only their names. 

\paragraph{Dataset}
We used an in-house dataset for mapping person names to World countries extracted from Wikipedia.\footnote{Paper is under revision.}

\subsection{Ethics and NLP: Factuality, Disinformation and Harmful content detection}    
\label{ssec:factuality_disinformation}

\subsubsection{Offensive Language Detection} 
The use of offensive language in social media has became a major problem, which can lead to real-world violence~\cite{husain2021survey,sap2019risk}. This literature for offensive language detection mainly focused on social media content and addressing for variety of languages. The task is mainly defined as whether the content (e.g., text, image, or multimodal) is offensive or not \cite{chowdhury2020multi}. 

\paragraph{Dataset}
For this task, we used the dataset from the SemEval-2020 Task 12 (OffensEval 2020) ~\citep{zampieri2020semeval}, which consists of 10,000 tweets, collected from a set of 660k Arabic tweets containing the vocative particle (``yA'' – O) from April 15 to May 6, 2019.

\subsubsection{Hate Speech Detection} 
\citet{davidson2017automated} defined hate speech as ``as language that is used to expresses hatred towards a targeted group or is intended to be derogatory, to humiliate, or to insult the members of the group''. The literature for hate speech detection defined the task as detecting hate vs. non-hate from different types of content such as text, image and multimodal \cite{schmidt2017survey,NEURIPS2020_1b84c4ce,gomez2020exploring}.


\paragraph{Dataset}
For this task, we also used the OSACT 2020 dataset ~\citep{mubarak-etal-2020-overview}, which consists of 10,000 tweets with annotated label hate-speech, not-hate-speech. 

\subsubsection{Adult Content Detection} 
Identifying this type of content is important for social media platforms to make a safe place for users. Especially this type of content poses a serious threat to other vulnerable groups (e.g., younger age groups). The task typically involves detecting and identifying whether the textual content contains sensitive/adult content or account that share such content. 

\paragraph{Dataset}
We used the dataset discussed in \citep{mubarak2021adult}, which contains 10,000 tweets collected by first identifying Twitter accounts that post adult content. Tweets are manually annotated as adult and not-adult.


\subsubsection{Spam Detection}
Spam content in social media includes ads, malicious content, and any low-quality content \cite{ghanem2023contents}. Spam detection is another important problem as such content may often annoy and mislead the users \cite{gao2012towards}. 

\paragraph{Dataset} 
We used the dataset discussed in \cite{mubarak2020spam} for Arabic spam detection which contains 28K tweets manually labeled as spam and not-spam.

\subsubsection{Subjectivity Identification} 
A sentence is considered subjective when it is based on -- or influenced by -- personal feelings, tastes, or opinions. Otherwise, the sentence is considered objective \cite{Francesco21}. Given that the identification of subjectivity is subjective itself, therefore, it poses challenges in the annotation process by the annotator. The complexity lies due to the different levels of expertise by the annotators, different interpretations and their conscious and unconscious bias towards the content they annotate. The content can be text (e.g., sentence, article), image or multimodal content, consisting of opinionated, factual or non-factual content. The annotation typically has been done using two labels, objective (OBJ) and subjective (SUBJ). 

\paragraph{Dataset} The dataset consists of sentences curated from news articles. The dataset has been developed based on the existing AraFacts dataset \cite{ali2021arafacts} that contains claims verified by Arabic fact-checking websites, and each claim is associated with web pages propagating or negating the claim. The news articles are collected from different news media. News articles were automatically parsed, split into sentences and filtered poorly-formatted sentences using a rule-based approach. A portion of the dataset was released as part of Task 2 in the CLEF2023 CheckThat Lab ~\cite{barron2023clef}.



\subsubsection{Propaganda Detection} 
Propaganda can be defined as a form of communication that aims to influence the opinions or the actions of people towards a specific goal; this is achieved utilizing well-defined rhetorical and psychological devices \cite{dimitrov-etal-2021-detecting}. In different communication channels, propaganda (persuasion techniques) is conveyed through the use of diverse techniques~\cite{Miller}, which range from leveraging the emotions of the audience, such as using \textit{emotional technique} or logical fallacies such as \textit{straw man} (misrepresenting someone's opinion), hidden \textit{ad-hominem fallacies}, and \textit{red herring} (presenting irrelevant data). 

\paragraph{Dataset} The dataset used for this study consists of Arabic tweets ~\citep{alam2022overview} posted by different news media from Arab countries such as Al Arabiya and Sky News Arabia from UAE, Al Jazeera, and Al Sharq from Qatar, and from five international Arabic news sources Al-Hurra News, BBC Arabic, CNN Arabic, France 24, and Russia Today. The final annotated dataset consists of 930 tweets. \citet{alam2022overview} formulated the task as a multilabel and multiclass span level classification task. For this study, we used the multilabel setup.

\subsubsection{Check-worthiness Detection}
Fact-checking is a time-consuming and complex process, and it often takes effort to determine whether a claim is important to check, irrespective of its potential to be misleading or not. Check-worthiness detection is the first step and a critical component of fact-checking systems \cite{survey:2021:ai:fact-checkers} and the aim is to facilitate manual fact-checking efforts by prioritizing the claims for the fact-checkers. Research on check-worthiness includes check-worthiness detection/ranking from political speeches, debates, and social media posts \cite{clef-checkthat:2022:task1,clef-checkthat:2021:task1}. A check-worthy claim is usually defined by its importance to the public and journalists, and whether it can cause harm to an individual, organization, and/or society.

\paragraph{Dataset} 
For this study, we used the Arabic subset of the dataset released with Task 1A (Arabic) of the CLEF2022 CheckThat Lab~\citep{nakov2022overview}. The dataset consists of 4,121 annotated tweets. The Arabic tweets were collected using keywords related to COVID-19, vaccines, and politics. 

\subsubsection{Factuality Detection}
\label{sssec:factuality_disinformation:factuality_detect}
Fact-checking has emerged as an important research topic due to a large amount of fake news, rumors, and conspiracy theories that are spreading in different social media channels to manipulate people’s opinions or to influence the outcome of major events such as political elections~\citep{darwish2017seminar,baly2018integrating}. While fact-checking has largely been done by manual fact-checker due to the reliability, however, that does not scale well as the enormous amount of information shared online every day. Therefore, an automatic fact-checking system is important and it has been used for facilitating human fact-checker \cite{survey:2021:ai:fact-checkers}. The task typically involves assessing the level of factual correctness in a news article, media outlets, or social media posts. The content is generally judged to be of high, low, or mixed factual correctness, using seven-point Likert scale\footnote{\url{https://mediabiasfactcheck.com}}$^,$\footnote{\url{https://allsides.com}} or just binary labels \{yes, no\} \cite{baly-etal-2018-predicting,alam-etal-2021-fighting-covid}.

\subsection*{Datasets}

\noindent 
\textbf{News Articles} We used the dataset developed by \citet{baly-etal-2018-predicting} in which false claims are extracted from verify-sy\footnote{\url{http://www.verify-sy.com}} and true claims are extracted from \url{http://ara.reuters.com}. The dataset consists of 3,042 documents.

\noindent
\textbf{Tweets}
For the claim detection from tweets, we used the same dataset \cite{alam-etal-2021-fighting-covid} discussed in Section~\ref{sssec:claim_detection}. As mentioned earlier, this dataset was annotated using a multi-questions annotation schema in which one of the questions was ``does the tweet appear to contain false information?''. Based on the answer to this question factuality label of the tweet has been defined. The Arabic dataset contains a total of 4,966 tweets.

\subsubsection{Claim Detection}
\label{sssec:claim_detection}
Information shared in the mainstream and social media often contains misleading content. Claim detection has become an important problem in order to mitigate misinformation and disinformation in those media channels. A factual (verifiable) claim is a sentence claiming that something is true, and this can be verified using factually verifiable information such as statistics, specific examples, or personal testimony \citep{DBLP:journals/corr/abs-1809-08193}. Research on claim detection includes social media posts -- text modality \cite{alam-etal-2021-fighting-covid}, multimodality \cite{cheema-etal-2022-mm} and news \cite{reddy2022newsclaims}.

\subsubsection*{Datasets}
\label{data:claim_detection}
\paragraph{CT-CWT-22-Claim} We used the Arabic subset of the dataset released with Task 1B of the CLEF2022 CheckThat Lab~\citep{clef-checkthat:2022:task1}. The dataset has been annotated using a multi-question annotation schema \cite{alam2020call2arms}, which consists of tweets collected using COVID-19 related keywords. The dataset contains 6,214 tweets \cite{nakov2022overview}.

\paragraph{ANS}~\cite{khouja-2020-stance} This dataset consists of 4,547 true and false claims, which was developed based on Arabic News Texts (ANT) corpus. A sample of articles was modified to generate true and false claims using crowdsourcing. 


\subsubsection{Harmful Content Detection}
For the harmful content detection we adopted the task proposed in \cite{alam-etal-2021-fighting-covid,nakov2022overview} though the research on harmful content detection also include identifying or detecting offensive, hate-speech, cyberbullying, violence, racist, misogynistic and sexist content~\cite{ijcai2022p781,alam-etal-2022-survey-1}. For some of the those harmful content detection tasks we addressed them separately and discussed in the below sections. \citet{alam-etal-2021-fighting-covid,nakov2022overview} proposed this concept in the context of tweets. The idea was to detect whether the content of a tweet aims to, and can, negatively affect society as a whole, specific individuals, companies, products, or spread rumors about them.
The content intends to harm or \textit{weaponize the information}\footnote{The use of information as a weapon to spread misinformation and mislead people.}~\citep{broniatowski2018weaponized}. 

\paragraph{Dataset}
We used the Arabic dataset proposed in \cite{nakov2022overview}, which consists of a total of 6,155 annotated tweets.

\subsubsection{Attention-worthiness Detection}
In social media most often people tweet by blaming authorities, providing advice, and/or call for action. It might be important for the policy makers to respond to those posts. The purpose of this task is to categorize such information into one of the following categories: \textit{not interesting, not sure, harmfullness, other, blames authorities, contains advice, calls for action, discusses action taken, discusses cure, asks a question}. 


\paragraph*{Dataset}
For this task, we used a subset of the dataset Task 1D of the CLEF2022 CheckThat Lab~\citep{clef-checkthat:2022:task1}, which contains 6,140 annotated tweets. 

\subsection{Semantic textual similarity} 

\subsubsection{Textual Similarity} Semantic textual similarity is a measure used to determine if two sentences are semantically equivalent. The task involves generating numerical similarity scores for pairs of sentences, with performance evaluated based on the Pearson correlation between machine-generated scores and human judgments~\cite{cer-etal-2017-semeval}. Two tasks were conducted to gauge the similarity between 250 pairs of Arabic sentences, as well as Arabic-English sentence pairs.
\paragraph{Dataset}
We used SemEval-2017 Task 1 (Track 1: ar-ar and Track 2: ar-en) dataset \cite{cer-etal-2017-semeval}, which is a translated version (machine translation followed by post-editing by human) of SNLI dataset~\cite{bowman-etal-2015-large}. 

\subsubsection{Semantic Question Similarity} 
The idea of this task is to determine how similar two questions are in terms of their meaning. 

\paragraph{Dataset}
We used Mawdoo3 Q2Q dataset (NSURL-2019 task 8: Semantic question similarity in Arabic), which consists of 15,712 annotated pairs of questions. Each pair is labeled as \textit{no semantic similarity (0)} or \textit{semantically similar (1)}~\cite{seelawi2019nsurl}.


\subsubsection{Natural Language Inference (NLI)} The XNLI task, known as Cross-lingual Natural Language Inference \cite{conneau2018xnli}, is a widely used benchmark in the field of natural language processing (NLP). It involves determining the logical relationship between pairs of sentences written in different languages. Specifically, the task requires NLP models to determine whether a given hypothesis sentence is entailed, contradicted, or neutral in relation to a given premise sentence, across multiple languages. The XNLI task serves as a rigorous evaluation of the cross-lingual transfer capabilities of NLP models, assessing their ability to understand and reason in different languages within a multilingual context.
\paragraph{Dataset}
The dataset we used for this study is the translated version of Arabic from XNLI corpus~\cite{conneau2018xnli}. For the annotation, 250 English sentences were selected from ten different sources and then asked the annotators to produce three hypotheses per sentence premise. The resulting premises and hypotheses are then translated into 15 languages and we used the Arabic version for this study.  

\subsection{Question Answering (QA)} 
This task involves answering questions in Arabic based on a given text\footnote{This task is also referred to as machine reading comprehension where the model is tested on its ability to extract answers from the given text}. For this task, we use four different datasets consisting of (passage, question, and answer) pairs.
\subsection*{Datasets}
\paragraph{ARCD} consists of 1,395 Arabic MSA questions posed by crowd-sourced workers along with the text segments from Arabic Wikipedia. We use the test set only for our evaluation. The test set consists of 78 articles, 234 paragraphs, and 702 questions \citep{mozannar2019neural}.

\noindent
\paragraph{MLQA} comprises multilingual question-answer instances in 7 languages, \emph{English}, \emph{Arabic}, \emph{Simplified Chinese}, \emph{Hindi}, \emph{German}, \emph{Vietnamese} and \emph{Spanish}. We used the Arabic QA pairs from this dataset, which consist of 2389 articles, 4646 paragraphs, and 5335 questions \citep{lewis2019mlqa}.

\noindent
\paragraph{TyDi QA} comprises 11 languages with 204K question-answer pairs. We used the data provided for the \emph{Gold Passage task} in which a passage that contains the answer is provided and the task is to predict the span that contains the answer. We used the Arabic split of the data which contains 921 articles, 921 paragraphs and 921 questions \citep{artetxe2019cross}.

\noindent
\paragraph{XQuAD} comprises 240 paragraphs and 1190 question-answers pairs from the development set of SQuAD v1.1 with their professional translations into ten languages. \emph{Hindi}, \emph{Turkish}, \emph{Arabic}, \emph{Vietnamese}, \emph{Thai}, \emph{German}, \emph{Greek}, \emph{Russian}, \emph{Spanish} and \emph{Chinese}. We use the the Arabic split of the data which consists of 48 articles, 240 paragraphs, and 1190 questions \citep{artetxe2019cross}.
We used the sQuad version of all datasets along with the official squad evaluation script.

\subsection{Speech Processing}
For this study, we address the speech modalities in the context of large foundation models, and we evaluate the following two tasks in this edition: (\textit{i}) automatic speech recognition (ASR); and (\textit{ii}) text to speech (TTS) models. In future, we will scale the speech benchmark with speech translation (ST) and spoken Arabic dialect identification spoken (ADI).

\begin{table*}[] 
\centering
\setlength{\tabcolsep}{6pt}
\resizebox{0.75\textwidth}{!}{
\begin{tabular}{llll}
\toprule
\textbf{Dataset} & \textbf{Task} & \textbf{Domain} & \textbf{Size} \\ \midrule
MGB2  & ASR & Broadcast (MSA) & 9.57 hrs \\
MGB3  & ASR & Broadcast (EGY) & 5.78 hrs \\
MGB5  & ASR & Broadcast (MOR) & 1.40 hrs \\
QASR.CS & ASR & Broadcast (Mixed) $\rightarrow$ Code-switching & 5.90 hrs \\
DACS  & ASR & Broadcast (MSA-EGY) $\rightarrow$ Code-switching & 1.50 hrs \\
ESCWA.CS & ASR & Meeting (Mixed DA - ENG) $\rightarrow$ Code-switching & 2.80 hrs \\
CallHome & ASR & Telephony (EGY) & 20 phone conversations \\
In-house & TTS & Mixed Topics (education, health, etc)  & 20 sentences \\
\bottomrule
\end{tabular}
}
\caption{Summary on test sets and their sizes used in evaluation for the speech processing tasks.}
\label{tab:datasets-speech}
\end{table*}


\subsubsection{Speech Recognition}
\label{ssec:asr}
The primary objective of an ASR system is to transform spoken language into written text. The task itself is challenging due to the presence of variability in human speech, which can be affected by factors such as accent, speaking style, code-switching, environmental factors like channels, and background noise among others. Furthermore, the presence of language-related challenges, including complex morphology, unstandardized orthography, and a wide array of dialects as a primary mode of communication, adds a layer of complexity to the task.
Therefore to properly benchmark Arabic ASR, we covered a wide range of domains encapsulating different speaking styles, dialects, and environments. For our study, we considered broadcast news, telephony, and meeting data for MSA, Egyptian, Moroccan Arabic, etc., in both monolingual and code-switching setups. 



\subsection*{Datasets}
\paragraph{MGB2} consists of $9.57$ hours of multi-dialect speech data that was collected from Aljazeera TV programs and manually transcribed. The data consists of a mix of Modern Standard Arabic (MSA) and various dialects, including Egyptian, Levantine, Gulf, and North African~\citep{ali2016mgb}.\footnote{\url{https://arabicspeech.org/mgb2}}

\paragraph{MGB3} is a collection of $5.78$ hours of multi-genre speech data in Egyptian dialect. The data was collected from YouTube videos and manually transcribed~\citep{ali2017speech}.\footnote{\url{https://arabicspeech.org/mgb3}}

\paragraph{MGB5} is a collection of $1.4$ hours of speech data in Moroccan dialect. The data was collected from YouTube videos and manually transcribed~\citep{ali2019mgb}.\footnote{\url{https://arabicspeech.org/mgb5}}

\paragraph{ESCWA.CS} is a collection of $2.8$ hours of speech code-switching corpus collected over two days of meetings of the United Nations Economic and Social Commission for West Asia (ESCWA) in 2019~\citep{chowdhury2021towards}.\footnote{\url{https://arabicspeech.org/escwa}}

\paragraph{QASR.CS} is a collection of $5.9$ hours of code-switching extracted from the Arabic broadcast news data (QASR) to test the system for code-switching. The dataset also includes some instances where the switch is between Arabic and French, however, this type of instance are very rare occurrence~\citep{mubarak2021qasr}.\footnote{\url{https://arabicspeech.org/qasr}}

\paragraph{DACS} is a collection of $\approx 1.5$ hours of broadcast speech designed to evaluate the performance of ASR for code-switching between MSA to Egyptian dialect and vice versa~\cite{chowdhury2020effects}.\footnote{\url{https://github.com/qcri/Arabic_speech_code_switching}}  

\paragraph{CallHome Egyptian} is a speech corpus of telephone conversations between native speakers of Egyptian Arabic. It consists of 20 unscripted telephone conversations, each of which lasts between 5-30 minutes~\citep{kumar-etal-2014-translations}.\footnote{\url{https://catalog.ldc.upenn.edu/LDC97S45}}

\subsubsection{Text to Speech}
Speech Synthesis a.k.a text to speech (TTS) helps users to get the written output easier and in some cases faster. Most state-of-the-art end-to-end TTS systems comprise three modules: text front-end, acoustic model, and vocoder. However, there is ongoing research to combine acoustic models and vocoder in a single neural network. Text front-end module normalizes input text by converting digits, symbols,  abbreviations, and acronyms  into full words, processing words with special sounds, borrowed words, etc. This task is challenging 
in Arabic due to missing  diacritics in modern texts as explained in \ref{ssec:diacritization}. Therefore, the Arabic front-end part of the TTS is responsible for restoring the missing diacritics and text normalization.

\paragraph{Dataset} For MSA TTS, we create the first public test dataset, which comprises 30 sentences covering different  topics such as psychology, education, health, etc. 
The average length for each sentence is 8 words. This data is used for objective and subjective evaluation for Arabic TTS.

\section{Model Parameters}
\label{sec:appx-model-parameters}

\subsection{NLP Models}
\label{sec:appx-nlp-model-parameters}
We used gpt-3.5-turbo-0301 and gpt-4-0314 versions for our tasks. In addition we used Bloomz 176B 8-bit version and Jais-13b chat version. 

\subsection{Speech Models}
\label{sec:appx-speech-model-parameters}

In Table \ref{tab:asr_models}, we provide the details of the speech model parameters. 

\begin{table}[ht]
\scalebox{0.9}{
\begin{tabular}{l|cccc}
\toprule
\multicolumn{1}{c|}{Model} & Layers & Width & Heads & Parameters \\ \hline
W.Small & 12 & 768 & 12 & 244M \\
W.Medium & 24 & 1024 & 16 & 769M \\
W.Large-v2 & 32 & 1280 & 20 & 1550m \\
USM & 32 & 1526 & 16 & 2B \\
\bottomrule
\end{tabular}}
\caption{Model parameters and architecture for Large pretrained ASRs. W. stands for Open.AI's Whisper \cite{radford2022robust} and USM is Universal Speech Model from Google \cite{zhang2023google} }
\label{tab:asr_models}
\end{table}

\section{Experiments and Results: Extended Details}
\label{sec:extended_results}
In this section, we provide extended versions of the results reported earlier in the paper.

\subsection{Random Baseline}
\label{ssec:appx_random_baseline}
For different tasks, we used different approaches to compute random baseline, as discussed below. 

\begin{itemize}
    \item \textbf{Segmentation:} We first randomly decide how many segments a token should have (between 0, 1 and 2), and then randomly split the characters of that token into the chosen number of segments.
    \item \textbf{Lemmatization:}  We first randomly decide the length of the lemma, and then randomly divide the remaining length between a prefix and suffix.
    \item \textbf{Diacritization:} We randomly choose between 9 choices for every character (8 diacritics and 1 choice for no diacritic).
    \item \textbf{QA:} Randomly select a span of tokens from the given context of each question.
    \item \textbf{Others (Multiclass and multilabel classification tasks):} For multiclass classification, we randomly assign a label to the test instance, with label selection based on the labels from the training set. For multilabel classification, which requires assigning multiple labels from a predefined set, both the number of labels and their selection were random, and these were assigned to the test instance.  
\end{itemize}

\subsection{Extended Few-shot Results}
\label{ssec:appx_extended_few_shot_results}

\begin{figure}[h]
    \centering
\includegraphics[width=0.9\columnwidth]{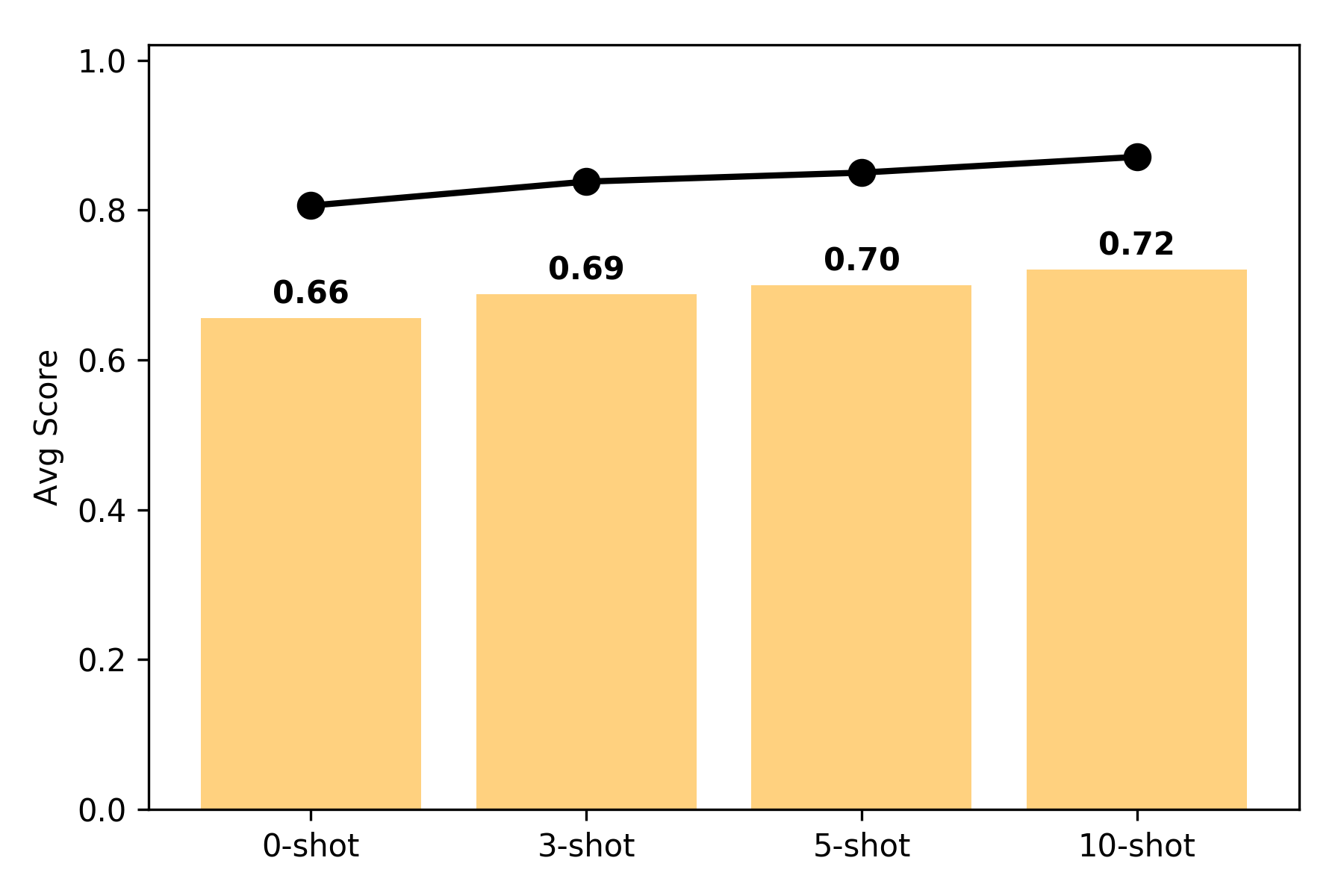}
    \caption{An average performance comparison (over seven tasks) of different few-shot experiments using GPT-4.}.
    \vspace{-0.4cm}
    \label{fig:few_shot_extended_exp}
    \vspace{-0.4cm}
\end{figure}

We conducted experiments using GPT-4 by incrementally increasing the number of shots. For this purpose, we chose one task from each of the seven groups listed in Table \ref{tab:exp_results_nlp_tasks} in the paper. We tested the models using 3, 5, and 10 shots. For each task, we observed a general trend of increasing performance, with the exception of the gender task. On average, performance improved from 0.656 in the 0-shot setting to 0.721 in the 10-shot setting. The results are presented in Table \ref{tab:few_shot_extended}. To provide a clear overview of the comparison across different few-shot scenarios, we present the average performance in Figure \ref{fig:few_shot_extended_exp}.


\subsection{Native Language Prompts}
\label{ssec:appx_native_language_prompts}

We have conducted experiments using Arabic prompts for the \textit{seven selected tasks}. The Arabic prompts were created by native Arabic speakers. The results are reported in Table \ref{tab:prompt_native_english}. Using the Arabic prompts, three out of the seven tasks outperformed their counterparts that used English prompts, two underperformed, and one showed equivalent performance. This finding partially supports the findings reported by \citet{ahuja2023mega}, which states that ``the monolingual prompting setup outperforms the cross-lingual prompting strategy''. However, they also report that using Davinci-003, the English prompts yield better results than their translated version in the native language.

\begin{table}[h]
\centering
\setlength{\tabcolsep}{3pt}
\resizebox{0.8\columnwidth}{!}{%
\begin{tabular}{@{}llrr@{}}
\toprule
\multicolumn{1}{c}{\textbf{Task Name}} & \multicolumn{1}{c}{\textbf{Metric}} & \multicolumn{1}{c}{\textbf{English}} & \multicolumn{1}{c}{\textbf{Arabic}} \\ \midrule
NER & Macro-F1 & 0.355 & 0.350 \\
Sentiment & Macro-F1 & 0.569 & 0.547 \\
News Cat. & Macro-F1 & 0.667 & 0.739 \\
Gender & Macro-F1 & 0.868 & 0.892 \\
Subjectivity & Macro-F1 & 0.677 & 0.725 \\
XNLI (Arabic) & Acc & 0.753 & 0.740 \\
QA & F1 (exact match) & 0.705 & 0.654 \\ \midrule
\textbf{Average} & \textbf{} & \textbf{0.656} & \textbf{0.664} \\ \bottomrule
\end{tabular}%
}
\caption{Results from GPT-4 using zero-shot prompts in both English and native languages.}
\label{tab:prompt_native_english}
\end{table}

\subsection{Semantic vs. Syntactic Task Differences}
\label{ssec:appx_sem_sync_task_diff}
We computed the performance difference between POS and MT, as shown in Table \ref{tab:semantic_syntactic_task}. The gap between SOTA and the three LLMs for POS (a syntactic task) is considerably larger than for MT (a semantic task). Moreover, the performance gap is much lower for semantic tasks compared to syntactic tasks, on average, across the three LLMs, as depicted in Table \ref{tab:semantic_syntactic_task}. This implies that these models might be better equipped to encode and express semantic information than to handle specific syntactic phenomena in their inputs.

\begin{table}[h]
\centering
\setlength{\tabcolsep}{3pt}
\resizebox{0.9\columnwidth}{!}{%
\begin{tabular}{@{}lrrrr@{}}
\toprule
\multicolumn{1}{c}{\textbf{}} & \multicolumn{1}{c}{\textbf{BLOOMZ}} & \multicolumn{1}{c}{\textbf{GPT-3.5}} & \multicolumn{1}{c}{\textbf{GPT-4}} & \multicolumn{1}{c}{\textbf{SOTA}} \\ \midrule
\multicolumn{5}{c}{\textbf{Semantic}} \\ \midrule
MT & \multicolumn{1}{r}{19.38} & 24.09 & 23.57 & 24.58 \\
Semantics (STS, XNLI) & \multicolumn{1}{r}{0.615} & 0.733 & 0.827 & 0.794 \\  \midrule
\multicolumn{5}{c}{\textbf{Syntactic}} \\  \midrule
POS & - & 0.154 & 0.464 & 0.844 \\
Parsing & - & 0.239 & 0.504 & 0.796 \\ \bottomrule
\end{tabular}
}
\caption{Average performance difference between semantic and syntactic tasks.}
\label{tab:semantic_syntactic_task}
\end{table}

\subsection{Performance Comparison with Jais}
\label{ssec:app_performance_jais}
Jais, as discussed in \cite{sengupta2023jais}, is an Arabic-focused model trained on English, Arabic, and programming code. To evaluate the Jais model, we employed the Jais-13b-chat variant and selected seven datasets corresponding to tasks outlined in Table \ref{tab:performance_jais}. For consistent output, we set the temperature parameter to zero and conducted the experiments in a zero-shot setting. The results presented in the table indicate that, on average, the performance of the Jais model surpasses that of random and BLOOMZ models. However, its performance falls below that of the models developed by OpenAI. For QA task, the performance of Jais is 4\% better than GPT-3.5. 

\begin{table*}[]
\centering
\scalebox{0.7}{
\begin{tabular}{@{}lllllllll@{}}
\toprule
\textbf{Task Name} & \textbf{Dataset} & \textbf{Metric} & \textbf{Random} & \textbf{BLOOMZ} & \textbf{Jais-13B-chat} & \textbf{GPT-3.5} & \textbf{GPT-4} & \textbf{SOTA} \\ \midrule
Sarcasm	& ArSarcasm	& F1 (POS)	& 0.240& 	0.286 &	0.288 &	0.465 &	0.400	& 0.46 \cite{farha2020arabic} \\
Sentiment & ArSAS & M-F1 & 0.222 & 0.251 & 0.304 & 0.550 & 0.569 & \textbf{0.758} \cite{hassan2021asad}\\
News Cat. & ASND & M-F1 & 0.048 & 0.371 &  0.195 & 0.512 & 0.667 & \textbf{0.770} \cite{chowdhury2020improving}\\
Gender & Arap-Tweet & M-F1 & 0.521 & {\color[HTML]{000000} 0.532} & 0.674 & 0.883 & 0.868 & \textbf{0.821} \cite{mubarak2022arabgend}\\
Subjectivity & In-house & M-F1 & 0.496 & 0.428 & 0.572 & 0.670 & 0.677 & \textbf{0.730} (In-house)\\
XNLI (Arabic) & XNLI & Acc & 0.332 & 0.500 & 0.425 & 0.489 &  \textbf{0.753} & 0.713 \cite{artetxe2019cross}\\
QA & ARCD & F1/EM & 0.085 & 0.368 & 0.546 & 0.502 & \textbf{0.705} & 0.613 \cite{mozannar2019neural}\\ \midrule
\textbf{Avg} &  &  &  \textbf{0.278}	 & \textbf{0.391}	& \textbf{0.429}	& \textbf{0.582} &	\textbf{0.663} &\textbf{0.695} \\
\bottomrule
\end{tabular}
}
\caption{Zero-shot performance comparison across models, including Jais, for seven datasets associated with seven different tasks. EM: Exact Match, M-F1: Macro-F1. Best result per task is \textbf{boldfaced}.}
\label{tab:performance_jais}
\end{table*}

\subsection{Qualitative Observations}
\label{ssec:app_qualitative_observations}

\begin{itemize}
    \item For sequence tagging tasks such as segmentation, POS tagging, NER, the common errors are \emph{(i)} output shape (either higher or lower), \emph{(ii)} return response with missing tokens, \emph{(iii)} inserts additional tokens, \emph{(iv)} instead of responding the label in English it provided responses in Arabic. Such errors are reflected in the high performance gap between SOTA and LLMs for these tasks.
    \item For multilabel tasks such as propaganda detection, the model returned response with additional labels that were not in the predefined label set.
    \item Bloomz Model: for syntactic tasks (e.g., segmentation, lemmatization, diacritization, POS, NER ), BLOOMZ consistently failed to produce any desired output, which might be that it does not understand the task at all. As for the diacritization task: It does not return any discretized content when instructed and answers by providing part of the input as output. This might be related to Arabic. However, it is worth looking into whether there is the same issue with other languages that use accented letters.
\end{itemize}

\subsection{Data Contamination Assessment}
\label{ssec:appx_data_contamination}
The presence of test data from standard downstream NLP tasks in the training dataset of pretrained LLMs' may effect the evaluations. It is important to have blind test-sets to reliably assert that the models are not merely memorizing data patterns but have truly acquired the ability to generalize. Identifying whether the data has been contaminated or not is a challenging problem. In our study, we have used the dataset that has been released after September 2021, which is a cut-off date for OpenAI's GPT models.\footnote{\url{https://platform.openai.com/docs/models/overview}} The tasks include CT–CWT–22 tasks (Checkworthy, Claim, Harmful content, and Attention-worthy) introduced in 2022. Consequently, for these specific tasks, the potential for data contamination is none.
Both GPT-3.5 and GPT-4 (in zero-shot and 3-shot scenarios) demonstrate results closely aligned with the state-of-the-art, mirroring trends seen in other 2021 test sets. In addition, the dataset for the subjectivity task is our in-house developed dataset, created at the end of 2022. 

To further validate whether evaluation datasets have been exposed to the LLMs, we assessed various datasets using the methodology outlined in \cite{golchin2023time}. 
It utilizes ``guided instruction'' as follows: a prompt consisting of the dataset name, partition type, and an initial reference instance, asking the LLM to complete it by providing second instance. An instance is flagged as contaminated if the LLM’s output either exactly or nearly matches with another instance. An example of an instruction is provided below.

\begin{lstlisting}
Instruction: You are provided with the first piece of an instance from the train split of the ArSAS dataset. Finish the second piece of the instance as exactly appeared in the dataset. Only rely on the original form of the instance in the dataset to finish the second piece.
Label: Negative
First Piece: {input instance}
Second Piece:
\end{lstlisting}

We applied this approach to GPT-4 across nine datasets associated with eight tasks: {\em(i)} Sentiment (ArSAS 2018), {\em(ii)} Emotion (SemEval-2018 Task 1, Arabic), {\em(iii)} Sarcasm (ArSarcasm-OSACT2020, ArSarcasm-v2-WANLP2021), {\em(iv)} News Category (ASND 2020), {\em(v)} Gender (Arap-Tweet 2022), {\em(vi)} Subjectivity (In-house 2022), {\em(vii)} XNLI 2020 (Arabic), and {\em(viii)} Question Answering (XQuAD 2019). For none of the nine datasets, corresponding to eight tasks, was GPT-4 able to produce any examples. Consequently, it is challenging to ascertain whether Arabic datasets for different tasks are included in the training data of ChatGPT. Thus, based on these experiments, we can conclude that the Arabic datasets for different tasks are not included in the training data of GPT models.

\subsection{Machine Translation (MT)}
\label{app:results:mt}
In Table \ref{tab:exp_results_full_mt_only}, we report detailed results for MT, considering both dialect and city levels.

\begin{table*}[!ht]
\centering
\setlength{\tabcolsep}{3pt}
\resizebox{0.7\textwidth}{!}{%
\begin{tabular}{@{}llllrrrrrr@{}}
\toprule
\multicolumn{1}{c}{\textbf{Dataset}} & \multicolumn{1}{c}{\textbf{Dialect}} & \multicolumn{1}{c}{\textbf{SC}} & \multicolumn{1}{c}{\textbf{City}} & \multicolumn{1}{c}{\textbf{\#Sent}} & \multicolumn{1}{c}{\textbf{BloomZ}} & \multicolumn{1}{c}{\textbf{Jais}}& \multicolumn{1}{c}{\textbf{\begin{tabular}[c]{@{}c@{}}Zero-shot \\ GPT-3.5\end{tabular}}} & \multicolumn{1}{c}{\textbf{\begin{tabular}[c]{@{}c@{}}Zero-shot \\ GPT-4\end{tabular}}} & \multicolumn{1}{c}{\textbf{SOTA}} \\ \midrule
APT & LEV & lv & - & 1000 & 11.38 & 13.13 & 18.55 & 17.77 & 21.9 \\
APT & Nile & eg & - & 1000 & 12.95 & 16.31 & 21.58 & 18.99 & 22.6 \\
MADAR & Gulf & iq & Baghdad & 2000 & 30.99 & 35.11 & 32.47 & 34.83 & 29.1 \\
MADAR & Gulf & iq & Basra & 2000 & 29.63 & 32.16 & 32.92 & 34.72 & 29 \\
MADAR & Gulf & iq & Mosul & 2000 & 29.17 & 32.49 & 30.82 & 35.32 & 31.3 \\
MADAR & Gulf & om & Muscat & 2000 & 39.91 & 39.17 & 39.37 & 39.9 & 39.5 \\
MADAR & Gulf & qa & Doha & 2000 & 31.1 & 33.26 & 33.6 & 33.62 & 29.3 \\
MADAR & Gulf & sa & Jeddah & 2000 & 40.37 & 39.51 & 42.62 & 42.69 & 29.4 \\
MADAR & Gulf & sa & Riyadh & 2000 & 27.73 & 31.1 & 32.51 & 33.71 & 40.7 \\
MADAR & Gulf & ye & Sana’a & 2000 & 29.79 & 32.7 & 32.48 & 34.63 & 31.4 \\
MADAR & LEV & jo & Amman & 2000 & 35.56 & 35.09 & 35.09 & 36.24 & 35.1 \\
MADAR & LEV & jo & Salt & 2000 & 34.54 & 32.76 & 35.78 & 37.54 & 34.9 \\
MADAR & LEV & lb & Beirut & 2000 & 24.01 & 28.43 & 26.14 & 28.95 & 23.7 \\
MADAR & LEV & ps & Jerusalem & 2000 & 34.02 & 34.39 & 35.22 & 35.5 & 33.6 \\
MADAR & LEV & sy & Aleppo & 2000 & 30.92 & 34.91 & 34.09 & 35.47 & 34.3 \\
MADAR & LEV & sy & Damascus & 2000 & 29.1 & 34.19 & 34.19 & 37.74 & 33.1 \\
MADAR & MGR & dz & Algiers & 2000 & 23.13 & 24.97 & 22.43 & 25.95 & 21.3 \\
MADAR & MGR & ly & Benghazi & 2000 & 25.41 & 29.07 & 26.99 & 30.12 & 32 \\
MADAR & MGR & ly & Tripoli & 2000 & 30.05 & 34.95 & 32.82 & 38.63 & 25.9 \\
MADAR & MGR & ma & Fes & 2000 & 23.73 & 28.87 & 22.53 & 26.15 & 29.9 \\
MADAR & MGR & ma & Rabat & 2000 & 31.02 & 35.86 & 31.95 & 34.71 & 23.1 \\
MADAR & MGR & tn & Sfax & 2000 & 15 & 20.78 & 15.93 & 20.74 & 13.8 \\
MADAR & MGR & tn & Tunis & 2000 & 16.79 & 18.77 & 14.69 & 18.51 & 16 \\
MADAR & MSA & ms & - & 2000 & 42.33 & 38.54 & 37.55 & 37.67 & 43.4 \\
MADAR & Nile & eg & Alexandria & 2000 & 29.24 & 32.96 & 32.05 & 32.46 & 38.3 \\
MADAR & Nile & eg & Aswan & 2000 & 39.97 & 39.68 & 41.77 & 42.42 & 30.4 \\
MADAR & Nile & eg & Cairo & 2000 & 32.79 & 32.15 & 32.77 & 32.69 & 32.9 \\
MADAR & Nile & sd & Khartoum & 2000 & 37.48 & 41.22 & 41.27 & 44.13 & 39 \\
MDC & LEV & jo & - & 1000 & 10.43 & 14.7 & 17.75 & 16.96 & 17.7 \\
MDC & LEV & ps & - & 1000 & 9.32 & 12.14& 15.72 & 14.22 & 15.3 \\
MDC & LEV & sy & - & 1000 & 10.24 & 15.83 & 18.66 & 16.96 & 19.9 \\
MDC & MGR & tn & - & 1000 & 8.28 & 12.8& 14.46 & 14.2 & 13.9 \\
MDC & MSA & ms & - & 1000 & 15.75 & 17.45 & 21.05 & 19.34 & 20.4 \\
Media & Gulf & om & - & 467 & 14.22 & 17.18 & 22.68 & 22.76 & 19.6 \\
Media & LEV & lb & - & 250 & 7.54 & 14.94& 17.65 & 16.65 & 16.8 \\
Media & MGR & ma & - & 526 & 4.87 & 11.05& 11.58 & 10.2 & 9.6 \\
Media & MSA & ms & - & 637 & 22.14 & 30.04 & 37.87 & 34.41 & 29.7 \\
Media & MSA & ms & - & 621 & 19.17 & 27.14 & 32.8 & 32.73 & 35.6 \\
Bible & MGR & ma & - & 600 & 16.34 & 20.34 & 16.16 & 15.14 & 28.8 \\
Bible & MGR & tn & - & 600 & 17.83 & 21.57 & 17.27 & 15.43 & 29.2 \\
Bible & MSA & ms & - & 600 & 24.37 & 25.94 & 23.96 & 18.38 & 33.2 \\
Bible & MSA & ms & - & 600 & 21.44 & 22.39 & 20.2 & 16.68 & 29.2 \\ \bottomrule
\end{tabular}%
}
\caption{Results (BLEU score) on machine translation for different datasets using zero-shot prompts. \#Sent. indicates number of sentences in test set. SOTA results are reported in \cite{sajjad2020arabench}.}
\label{tab:exp_results_full_mt_only}
\end{table*}

\section{Prompts}
\label{sec:appx-prompts}

The performance of the model is highly dependent on the prompting strategy. Designing the best prompts for each task is challenging and required several iterations. In many tasks, the output was not consistent for all instances of the datasets. 
For example, in many cases the model provides the desired labels, 
however, there are cases where the model output different kind of error messages: {\em(i)} it is trained only on English and cannot handle Arabic texts, {\em(ii)} the response was filtered due to the prompt triggering Azure OpenAI’s content management policy, {\em(iii)} it often provided extra tokens or swapped the tag (B-PER to PER-B). These resulted in an extra layer of post-processing and filtering of the evaluation dataset.  
Moreover, from our initial exploration, we noticed that, compared to language-specific (Arabic) prompts, English prompts (task-description) provide superior performance. Our underlying hypothesis is that with English task-description the input representations shift toward the English space that allows the model to process and understand the input better, giving better performance.\footnote{Note this observation aligns with other multilingual low-resource language studies.}

For the segmentation task, with our initial prompt, we realized that the output was not segmented based on linguistic information but rather more Byte-Pair Encoding (BPE) like encoding. Based on that prompt is further redesigned, which resulted in a better outcome. 

For factuality, disinformation, and harmful content detection tasks, the challenges were different from other tasks. One notable example is the propaganda detection task. The task requires determining whether a text snippet contains propagandistic language, and if it does, the model should detect which propaganda technique is used from a pre-defined list of techniques. Even with our best efforts to design the prompt for this task, the model still produced very unexpected responses, sometimes incomplete names of propaganda techniques, or even techniques not among the provided list. 

Another challenge with designing prompts for these tasks, is the issue of a task's subjectivity where providing a crisp-clear classification task definition to the model is not possible. As an example, one of our tasks is to evaluate whether a tweet is offensive towards a person or an entity. In many instances, the model predicted tweets to be offensive, while in reality they were descriptive of the tweet's author mental or physical state, or they were just repeating common negative statements or Arabic proverbs not directed at anyone indicating the model's understanding of offensiveness is not inline of our definition.

In the following sections, we report a set of prompts we used for different tasks. However, this is not exhaustive and does not cover all prompts for all the different models and settings. We kindly refer the reader to our LLMeBench framework~\cite{dalvi-etal-2024-llmebench} to find a complete list. 

\subsection{Word Segmentation, Syntax and Information Extraction}
\label{ssec:appendix:word_egmentation}
\subsubsection*{Segmentation}


\begin{lstlisting}
A word can be composed of one root and one or multiple affixes. Segment the following sentence into its morphological constituents: {inputSentence}"+". The output format should be a list of tuples, where each tuple consists of a word from the input text and its segmented form joined by a + sign.
\end{lstlisting}
\subsubsection*{Named Entity Recognition\footnote{prompt was inspired by \cite{lai2023chatgpt}}}


\begin{lstlisting}
Task Description: You are working as a named entity recognition expert and your task is to label a given arabic text with named entity labels. Your task is to identify and label any named entities present in the text without any explanation. The named entity labels that you will be using are PER (person), LOC (location), ORG (organization), MISC (miscellaneous). You may encounter multi-word entities, so make sure to label each word of the entity with the appropriate prefix ('B' for first word entity, 'I' for any non-initial word entity). For words which are not part of any named entity, you should return 'O'. Note: Your output format should be a list of tuples, where each tuple consists of a word from the input text and its corresponding named entity label. Input: {inputSentence}
\end{lstlisting}

\subsubsection*{POS}
\begin{lstlisting}
These are the segmentation and POS tags for a sample sentence:
`\<فيلم جاذبية يتصدر ترشيحات جوائز الأكاديمية البريطانية
لفنون الفيلم والتلفزيون>`
`\<فيلم>`    `\<فيلم>`    NOUN
`\<جاذبية>`    `\<جاذبي +   ة>`    NOUN+NSUFF
`\<يتصدر>`    `\<يتصدر>`    V
`\<ترشيحات>`    `\<ترشيح + ات>`    NOUN+NSUFF
`\<جوائز>`    `\<جوائز>`    NOUN
`\<الأكاديمية>`   `\<ال + أكاديمي + ة>`     DET+NOUN+NSUFF
`\<البريطانية>`   `\<ال + بريطاني + ة>`     DET+ADJ+NSUFF
`\<لفنون>`    `\<ل + فنون>`    PREP+NOUN
`\<الفيلم>`    `\<ال + فيلم>`    DET+NOUN
`\<والتلفزيون>`    `\<و + ال + تلفزيون>`    CONJ+DET+NOUN

get the segmentation and POS tags for this sentence: {inputSentence}
\end{lstlisting}

\begin{lstlisting}
Assign POS tag to each morphological segment within each word. group the tags for each word with +: {inputSentence}"+". The output should be in the format: [{word: label}, {word: label}]
\end{lstlisting}

\begin{lstlisting}
Label the following sentence with its corresponding PENN Treebank POS Labels. 
sentence: {inputSentence}
labels:
\end{lstlisting}

\subsubsection*{Lemmatization}
\begin{lstlisting}
for every word in the following sentence, write only the lemmas without diacritics in separate lines without explanation:
{inputSentence}
\end{lstlisting}

\subsubsection*{Diacritization}
\begin{lstlisting}
Diacritize fully the following Arabic sentence: {inputSentence}
\end{lstlisting}

\begin{lstlisting}
Vowelized the following sentence: {inputSentence}. Words that can't be vowelized put them back as they were.
\end{lstlisting}

\subsubsection*{Parsing}
\begin{lstlisting}
Given the following features (in order: ID, Form, Lemma, CPostTag, POSTag, Features), predict the Head of each token in the following sentence, which is either a value of a related ID or 0. A value of zero means the token attaches to the virtual root node: {inputSentence}
\end{lstlisting}

\subsubsection*{Dialect Identification}
\begin{lstlisting}
Write only the country code of the Arabic country in which this sentence is written in its dialect without any explanation? Write only the country code in ISO 3166-1 alpha-2 format without explanation. Write 'MSA' if the sentence is written in Modern Standard Arabic.
sentence: {inputSentence}
code: 
\end{lstlisting}


\subsection{Sentiment, Stylistic and Emotion Analysis} 

\subsubsection*{Sentiment analysis} 
\begin{lstlisting}
Choose only one sentiment between: Positive, Negative, Neutral, or Mixed for this sentence:
sentence: {inputSentence}
label: 
\end{lstlisting}

\subsubsection*{Emotion detection} 
\begin{lstlisting}
Predict all the possible emotions in the following Arabic sentences without explanation and put them in a Python list. List of emotions are: anger, anticipation, disgust, fear, joy, love, optimism, pessimism, sadness, surprise, and trust
sentence: {inputSentence}
labels: 
\end{lstlisting}



\subsection{Demographic/Protected Attributes}
\subsubsection*{Gender}
\begin{lstlisting}
If the following person name can be considered as male, write 'm' without explnanation, and if it can be considered as female, write 'f' without explnanation.
person name: {inputSentence}
label: 
\end{lstlisting}
\subsubsection*{Location} 
\begin{lstlisting}
Map the following locations to one of the Arab countries. Write the country code in ISO 3166-1 alpha-2 format without explanation. If the country is outside Arab countries, write 'OTHERS', and if the location cannot be mapped to any country in the world, write 'UNK' without any explanation.
location: {inputSentence}
label:
\end{lstlisting}

\subsubsection*{Name Info}
\begin{lstlisting}
Predict the country of citizenship of the following person name. Write the country code in ISO 3166-1 alpha-2 format without explanation.
name: {inputSentence}
code: 
\end{lstlisting}

\subsection{Ethics and NLP: Factuality, Disinformation, Harmful content}
\subsubsection*{Offensive Language}
\begin{lstlisting}
If the following sentence is offensive, just write "OFF", otherwise, just write "NOT_OFF" without explanation:
sentence: {inputSentence}
label: 
\end{lstlisting}

\subsubsection*{Hate Speech}
\begin{lstlisting}
If the following sentence has hate speech, just write "HS", otherwise, just write "NOT_HS" without explanation:
sentence: {inputSentence}
label:
\end{lstlisting}

\subsubsection*{Adult Content}
\begin{lstlisting}
Classify the following Arabic sentence as adult language (the language used in adult advertisement and porno advertisement) or not adult language without illustruation. In case of adult language, just write "ADULT" without explaination, and in case of not adult language, just write "NOT_ADULT" without explanation.
text: {inputSentence}
label: 
\end{lstlisting}

\subsubsection*{Spam}
\begin{lstlisting}
If the following sentence can be classified as spam or contains an advertisemnt, write 'ADS' without explnanation, otherwise write 'NOTADS' without explanantion.
sentence: {inputSentence}
label:
\end{lstlisting}

\subsubsection*{Subjectivity}
\begin{lstlisting}
Classify the sentence as subjective or objective. Provide only label.
sentence: {inputSentence}
label:

\end{lstlisting}

\subsubsection*{Checkworthiness}
\begin{lstlisting}
Classify the sentence as checkworthy or not checkworthy. Provide only the label.
sentence: {inputSentence}
label:
\end{lstlisting}

\subsubsection*{Claim detection}
\begin{lstlisting}
Does this sentence contain a factual claim? Answer only by yes or no.
sentence: {inputSentence}
label:
\end{lstlisting}

\subsubsection*{Harmful content detection}
\begin{lstlisting}
Classify the following sentence as harmful or not harmful. Answer only by yes or no. Provide only label.
sentence: {inputSentence}
label:
\end{lstlisting}

\subsubsection*{Attention-worthy}
\begin{lstlisting}
Classify the sentence by whether it should get the attention of policymakers. Answer by yes or no. If the predicted label is yes then classify the sentence into one of the following categories: asks question, blame authorities, calls for action, Harmful, contains advice, discusses action taken, discusses cure, or other.
text: {input_sample}
label: 
\end{lstlisting}

\subsection{Semantics}

\subsection*{Semantic Textual Similarity}
\begin{lstlisting}
Given two sentences, produce a continuous valued similarity score on a scale from 0 to 5, with 0 indicating that the semantics of the sentences are completely independent and 5 indicating semantic equivalence. The output should be exactly in the form of a similarity score.
sentence 1: {inputSentence1}
sentence 2: {inputSentence2}
score: 
\end{lstlisting}

\subsection*{Natural Language Inference}
\begin{lstlisting}
You are provided with a premise and a hypothesis. Your task is to classify the hypothesis as true (entailment), false (contradiction), or unknown (neutral) based on the given premise. The output should be true, false or unknown. 
premise: {inputSentence1}
hypothesis: {inputSentence2}
output: 
\end{lstlisting}

\subsubsection*{Classification (Question Similarity)}
\begin{lstlisting}
Are the following two questions semantically similar? The output should be exactly either yes or no.
question 1: {inputQuestion1}
question 2: {inputQuestion2}
label:
\end{lstlisting}

\subsection{Question answering (QA)} 

\begin{lstlisting}
Your task is to answer questions in Arabic based on a given context.
Note: Your answers should be spans extracted from the given context without any illustrations. 
You don't need to provide a complete answer.
context:{context}                                                           
question:{question}
answer:
\end{lstlisting}


\begin{table*}[]
\centering
\resizebox{0.94\textwidth}{!}{
\begin{tabular}{lllllllllllll}
\toprule
\multicolumn{1}{c}{\textbf{Reference}} & \multicolumn{1}{c}{\textbf{\# tasks}} & \multicolumn{1}{c}{\textbf{\# datasets}} & \multicolumn{1}{c}{\textbf{\begin{tabular}[c]{@{}c@{}}Fine-tuned\\Models\end{tabular}}} & \multicolumn{1}{c}{\textbf{\begin{tabular}[c]{@{}c@{}}Zero-shot\\ GPT-3.5\end{tabular}}} & \multicolumn{1}{c}{\textbf{\begin{tabular}[c]{@{}c@{}}Few-shot\\ GPT-3.5\end{tabular}}} & \multicolumn{1}{c}{\textbf{\begin{tabular}[c]{@{}c@{}}Zero-shot\\ GPT-4\end{tabular}}} & \multicolumn{1}{c}{\textbf{\begin{tabular}[c]{@{}c@{}}Few-shot\\ GPT-4\end{tabular}}} & \multicolumn{1}{c}{\textbf{\begin{tabular}[c]{@{}c@{}}Zero-shot\\ BLOOMZ\end{tabular}}} & \multicolumn{1}{c}{\textbf{\begin{tabular}[c]{@{}c@{}}Zero-shot\\ Jais-13B-chat\end{tabular}}} & \multicolumn{1}{c}{\textbf{\begin{tabular}[c]{@{}c@{}}SOTA\\Comp.\end{tabular}}} & \multicolumn{1}{c}{\textbf{Modality}} \\ \midrule
AraBench~\cite{sajjad2020arabench} & 1 & 6 & \begin{tabular}[c]{@{}l@{}}Seq2Seq \\(transformer)\end{tabular} & \xmark & \xmark & \xmark & \xmark & \xmark & \xmark & \cmark & T, S \\
ARLUE~\cite{abdul2021arbert} & 13 & 42 & \begin{tabular}[c]{@{}l@{}}ARBERT, \\MARBERT\end{tabular} & \xmark & \xmark & \xmark & \xmark &  & \xmark &  \cmark & T \\
ALUE~\cite{seelawi2021alue} & 8 & 8 & AraBERT, mBERT & \xmark & \xmark & \xmark & \xmark & \xmark & \xmark & \cmark & T \\
ORCA~\cite{elmadany2022orca} & 29 & 60 & \begin{tabular}[c]{@{}l@{}}mBERT, ARBERT, \\ CamelBERT, \\ MARBERT\end{tabular} & \xmark & \xmark & \xmark & \xmark & \xmark & \xmark & \cmark & T \\ 
GPTAraEval~\cite{tawkat2023gptaraeval} & 44 & 60 & MARBERT, AraT5 & \cmark & \cmark & \cmark & \xmark & \cmark & \xmark & \xmark & T \\ \midrule
\textbf{LAraBench (Ours)} & 33 & 61 & \xmark & \cmark & \xmark & \cmark & \cmark & \cmark &  \cmark &  \cmark & T, S \\ \bottomrule
\end{tabular}
}
\caption{A comparison with prior studies. T: Text, S: Speech.}
\label{tab:prev_studies_comparison}
\end{table*}

\section{Post-processing} Post-processing was needed for almost all tasks in order to match gold labels, which include reformatting the output handling exceptions, missing values, and unexpected values. Much like NLP tasks, post-processing the transcription output from the speech models is an important step. We noticed that the performance of the Whisper models is highly dependent on the post-processing. As the models (Whisper family) are trained with massive dataset created by weak supervision, the output is quite noisy and needs extra care for post-processing. In this study, we opt for a simple post-processing pipeline so that the process is not overfitted to task-based data styles.

\section{Benchmarks on Arabic: Details}
\label{sec:appx_benchmark_on_arabic}
In this section, we discuss the work related to Arabic that has been conducted for benchmarking purposes.

\textit{\textbf{GPTAraEval}~\cite{tawkat2023gptaraeval}} is a large-scale automated and human evaluation of ChatGPT in zero- and few-shot settings, covering 44 distinct Arabic language understanding and generation tasks on 60 different datasets. The model is also compared to the open model BLOOMZ and two fine-tuned Arabic language models. Furthermore, comparison of ChatGPT and GPT-4’s performance on Modern Standard Arabic and Dialectal Arabic is conducted on a handful of tasks. It should be noted that the work only tested the models on  \textit{a sample of 200 points} from each of the evaluation test sets.

\textit{\textbf{ORCA}} ~\cite{elmadany2022orca}, a large-scale benchmark that incorporates 60 diverse datasets organized into seven comprehensive task clusters. This large-scale organization allows for a more in-depth and diverse analysis of model performance across a multitude of language tasks including but not limited to sentence classification, text classification, structured prediction, semantic similarity, natural language inference, question-answering, and word sense disambiguation.

\textit{\textbf{AraBench}} ~\cite{sajjad2020arabench} is an evaluation suite for dialectal Arabic-to-English machine translation. It offers a wide range of dialect categories including 4 coarse, 15 fine-grained, and 25 city-level dialects from various genres like media, chat, and travel. It also provides robust baselines that utilize different training methods like fine-tuning, back-translation, and data augmentation. 

The \textbf{\textit{ALUE}} ~\cite{seelawi2021alue} benchmark offers 8 curated tasks and private evaluation datasets, covering areas like emotion classification, hate speech, and fine-grained dialect identification. ArabicBERT tops the performance in 7 of these 8 tasks, with evaluations also including BERT variants with AraVec and FastText models.

\textbf{\textit{ARLUE}} ~\cite{abdul2021arbert} benchmark employs 42 datasets for six task clusters to evaluate multi-dialectal Arabic language understanding, featuring BERT and XLM model variants. Fine-tuned models utilizing ARLUE lead the performance in all six clusters.

As shown in Table~\ref{tab:prev_studies_comparison}, Our study provides a comprehensive evaluation platform that advances the current benchmarks by presenting 33 distinct tasks over 61 datasets, which is the most extensive task coverage among current benchmarks. Unlike the AraBench, which focuses exclusively on Arabic-to-English translation tasks, and ALUE and ARLUE, which have a narrower task focus or a lesser number of tasks, \arabench{} provides a broader scope of evaluation tasks. This benchmark encompasses a multitude of language tasks that are paramount to understanding the robustness and generalizability of language models. Furthermore, \arabench{} distinguishes itself by not only including text modality but also speech modality, thereby increasing the robustness and utility of our benchmark. Additionally, we successfully evaluated GPT-3.5, GPT-4, BLOOMZ, and Jais demonstrating its compatibility with cutting-edge language models.

Notably, the models employed in \arabench{} have displayed comparable performance with the SOTA models, attesting to its robustness and high standard of evaluation. While SOTA models generally outperform LLMs, our benchmark reveals that these LLMs can close the performance gap in certain tasks, particularly when increasing prompt complexity and transitioning from zero-shot to few-shot learning. This highlights LAraBench's utility not only as a tool for model evaluation but also as an instrumental platform for identifying tasks under which LLMs might be able to match or even surpass SOTA performance. This benchmark serves as a challenging testbed for future language models and contributes to the advancement of Arabic language understanding models.

\end{document}